\documentclass[a4paper,fleqn]{cas-dc}

\usepackage{tikz}
\usepackage{cleveref}

\usepackage[authoryear]{natbib}
\usepackage{xcolor}

\newcommand{\rev}[1]{\textcolor{black}{#1}}

\def\tsc#1{\csdef{#1}{\textsc{\lowercase{#1}}\xspace}}
\tsc{WGM}
\tsc{QE}
\tsc{EP}
\tsc{PMS}
\tsc{BEC}
\tsc{DE}

\begin{document}

\tolerance=999
\sloppy

\let\WriteBookmarks\relax
\def\floatpagepagefraction{1}
\def\textpagefraction{.001}

\newcommand{\etal}{\textit{et al.}}

\shortauthors{A. Lopes et~al.}

\shorttitle{CCNeXt: An Effective Self-Supervised Stereo Depth Estimation Approach}

\title{CCNeXt: An Effective Self-Supervised Stereo Depth Estimation Approach}                      



%
\author[1,2]{Alexandre Lopes}[type=author, orcid=0000-0002-3873-2835,
  auid=000,bioid=1]

\cormark[1]


\ead{a115968@dac.unicamp.br}

\credit{Conceptualization of this study, Methodology, Software}

\affiliation[1]{organization={Institute of Computing, University of Campinas},
    country={Brazil}}

\affiliation[2]{organization={Venturus - Innovation \& Technology},
    country={Brazil}}
\author[3]{Roberto Souza}[type=author, orcid=0000-0001-7824-5217,
  auid=000,bioid=2]

\author[1]{Helio Pedrini}[type=author, orcid=0000-0003-0125-630X,
  auid=000,bioid=3]

\affiliation[3]{organization={Department of Electrical and Computer Engineering, University of Calgary},
    country={Canada}}


\cortext[cor1]{Corresponding author}



\begin{abstract}
Depth Estimation plays a crucial role in recent applications in robotics, autonomous vehicles, and augmented reality. These scenarios commonly operate under constraints imposed by computational power. Stereo image pairs offer an effective solution for depth estimation since it only needs to estimate the disparity of pixels in image pairs to determine the depth in a known rectified system. Due to the difficulty in acquiring reliable ground-truth depth data across diverse scenarios, self-supervised techniques emerge as a solution, particularly when large unlabeled datasets are available. We propose a novel self-supervised convolutional approach that outperforms existing state-of-the-art Convolutional Neural Networks (CNNs) and Vision Transformers (ViTs) while balancing computational cost. The proposed CCNeXt architecture employs a modern CNN feature extractor with a novel windowed epipolar cross-attention module in the encoder, complemented by a comprehensive redesign of the depth estimation decoder. Our experiments demonstrate that CCNeXt achieves competitive metrics on the KITTI Eigen Split test data while being 10.18$\times$ faster than the current best model and achieves state-of-the-art results in all metrics in the KITTI Eigen Split Improved Ground Truth and Driving Stereo datasets when compared to recently proposed techniques. To ensure complete reproducibility, our project is accessible at 
\href{https://github.com/alelopes/CCNext}{\texttt{https://github.com/alelopes/CCNext}}.
\end{abstract}








\maketitle

\section{Introduction}
\label{sec:intro}

Depth information is critical for many computer vision and image analysis applications. It has been applied to tasks such as synthetic object insertion in computer graphics~\citep{luo2020consistent}, robot-assisted surgery~\citep{stoyanov2010real}, autonomous driving~\citep{5940562}, and smart glasses~\citep{wang2023practical}. These applications typically encounter constraints related to computing resources.

Depth can be estimated using a monocular camera~\citep{bhat2023zoedepth}, stereo camera~\citep{fang2023es3net}, or a multi-camera system~\citep{wei2023surrounddepth}. Monocular depth estimation techniques involve learning meter or relative depth information that can be extracted from individual images, a process that necessitates supervised training~\citep{guizilini20203d} or estimating additional parameters such as camera parameters for self-supervised techniques using monocular videos~\citep{guizilini2022multi,godard2019digging}. In contrast, stereo systems can solve depth estimation by calculating the disparity of pixels in a pair of cameras. For rectified images, knowing the intrinsic and extrinsic camera parameters is sufficient to transform depth estimation into a stereo matching problem. Self-supervised methods can also be applied to stereo systems, where the primary objective is to establish correspondences between left and right images without relying on ground-truth data. Since the camera system is known, the main advantage of a stereo system is that it obviates the necessity of estimating the inter-frame camera poses. This is particularly advantageous in complex scenarios involving object motion~\citep{godard2019digging}.

\tikzset{every picture/.style={line width=0.75pt}} 
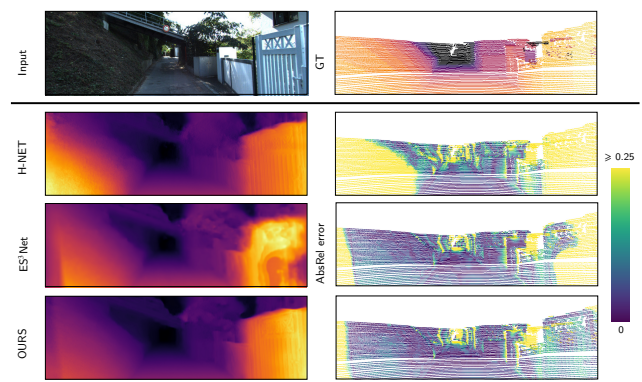
\begin{figure}[t]
\centering
\begin{minipage}{\columnwidth}
\resizebox{1\textwidth}{!}{

\begin{tikzpicture}[x=0.75pt,y=0.75pt,yscale=-1,xscale=1]

\draw (477.48,54.06) node  {\includegraphics[width=206.14pt,height=65.11pt]{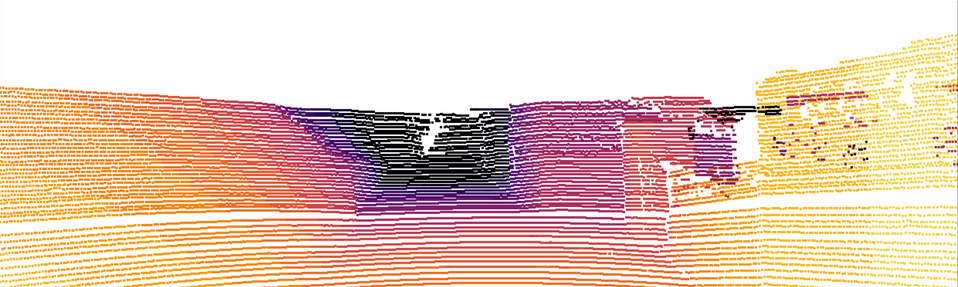}};
\draw (173.76,54.67) node  {\includegraphics[width=207.12pt,height=66.02pt]{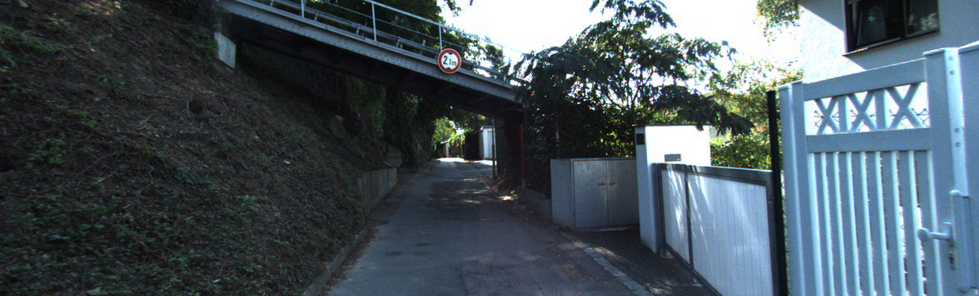}};
\draw (172.99,160.3) node  {\includegraphics[width=205.96pt,height=65.41pt]{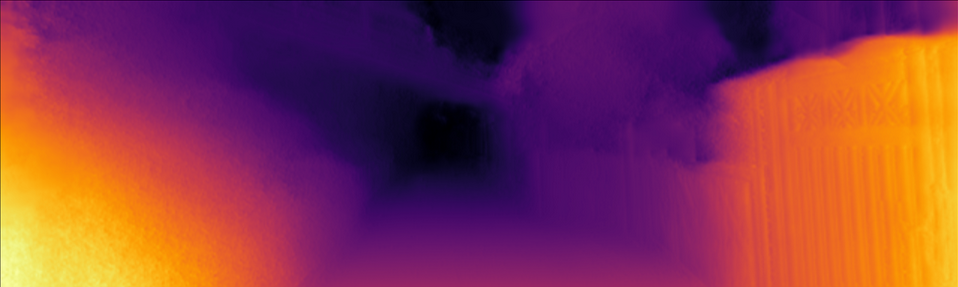}};
\draw (173,256.63) node  {\includegraphics[width=205.98pt,height=66.06pt]{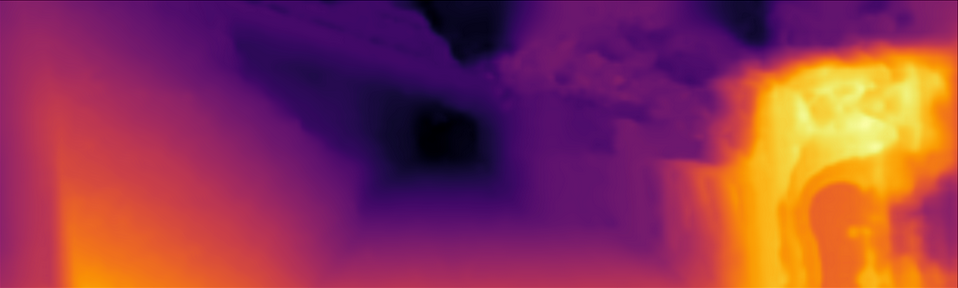}};
\draw (173,354.01) node  {\includegraphics[width=205.98pt,height=65.99pt]{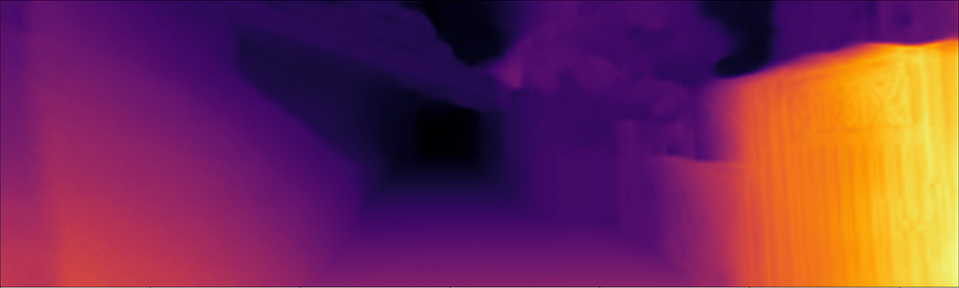}};
\draw (477.96,160.31) node  {\includegraphics[width=206.26pt,height=65.44pt]{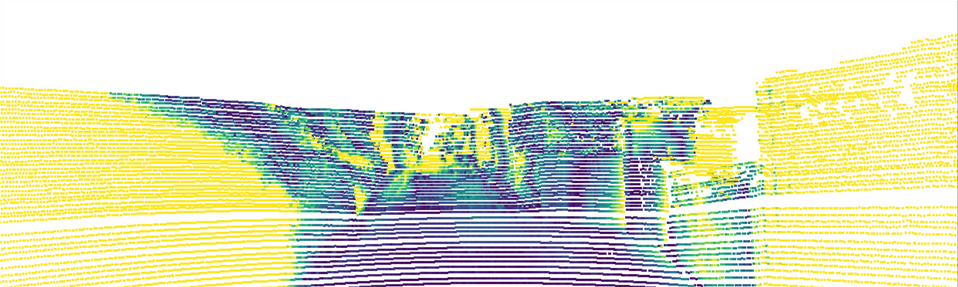}};
\draw (478.6,255.45) node  {\includegraphics[width=206.26pt,height=65.44pt]{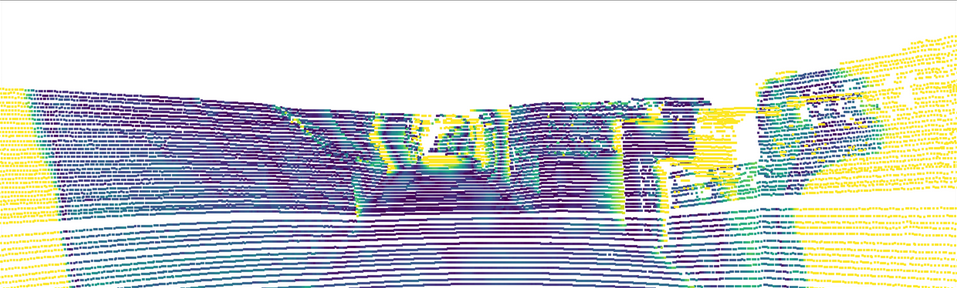}};
\draw (478.6,353.36) node  {\includegraphics[width=206.26pt,height=65.44pt]{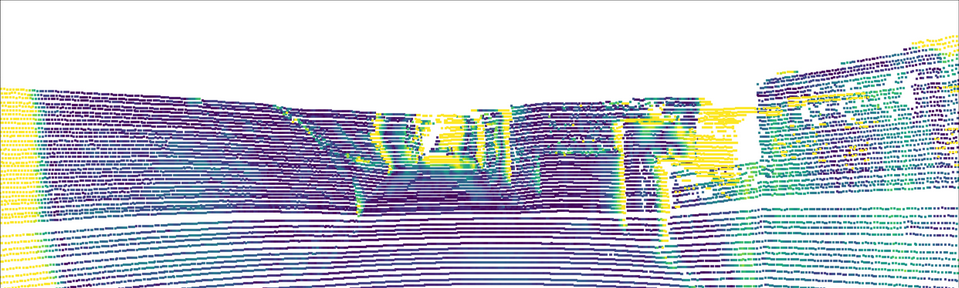}};
\draw (639.52,256.05) node [rotate=-270] {\includegraphics[width=120.14pt,height=14.22pt]{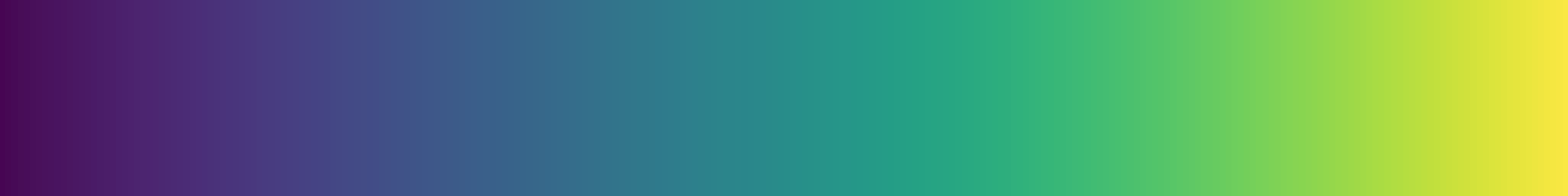}};
\draw   (340.46,116.69) -- (615.47,116.69) -- (615.47,203.94) -- (340.46,203.94) -- cycle ;
\draw   (341.1,211.83) -- (616.11,211.83) -- (616.11,299.08) -- (341.1,299.08) -- cycle ;
\draw   (341.1,309.73) -- (616.11,309.73) -- (616.11,396.98) -- (341.1,396.98) -- cycle ;
\draw   (35.68,310.02) -- (310.69,310.02) -- (310.69,397.27) -- (35.68,397.27) -- cycle ;
\draw   (35.68,212.59) -- (310.69,212.59) -- (310.69,299.84) -- (35.68,299.84) -- cycle ;
\draw   (35.68,116.69) -- (310.69,116.69) -- (310.69,203.94) -- (35.68,203.94) -- cycle ;
\draw   (35.68,10.65) -- (310.69,10.65) -- (310.69,97.91) -- (35.68,97.91) -- cycle ;
\draw   (340.06,10.65) -- (615.07,10.65) -- (615.07,97.91) -- (340.06,97.91) -- cycle ;
\draw   (0,106.65) -- (659.5,106.65) -- (659.5,107.65) -- (0,107.65) -- cycle ;

\draw (318.31,298.45) node [anchor=north west][inner sep=0.75pt]  [rotate=-270] [align=left] {AbsRel error};
\draw (5.68,79.85) node [anchor=north west][inner sep=0.75pt]  [rotate=-270] [align=left] {Input};
\draw (318.31,73.32) node [anchor=north west][inner sep=0.75pt]  [rotate=-270] [align=left] {GT};
\draw (5.68,184.3) node [anchor=north west][inner sep=0.75pt]  [rotate=-270] [align=left] {H-NET};
\draw (5.68,286.52) node [anchor=north west][inner sep=0.75pt]  [rotate=-270] [align=left] {ES$^3$Net};
\draw (5.68,377.96) node [anchor=north west][inner sep=0.75pt]  [rotate=-270] [align=left] {OURS};
\draw (636.52,339.84) node [anchor=north west][inner sep=0.75pt]  [font=\footnotesize] [align=left] {0};
\draw (619.2,158.49) node [anchor=north west][inner sep=0.75pt]  [font=\footnotesize] [align=left] {\begin{minipage}[lt]{27.72pt}\setlength\topsep{0pt}
\begin{center}
$\geqslant $ 0.25
\end{center}

\end{minipage}};

\end{tikzpicture}

}
\end{minipage}
  \caption{\textbf{Above the line}: Input image and depth map ground truth (GT).  \textbf{Below the line}: Depth estimations from H-Net~\citep{huang2022h}, ES$^3$Net~\citep{fang2023es3net} and our method on the left. The AbsRel metric error is shown on a map on the right for all estimations when compared to the ground truth. For the AbsRel error map, the darker/bluer the point is, the lower the error.}
  
  \label{fig:1}

\end{figure}

CNNs have been successfully applied to depth estimation and stereo matching~\citep{wang2021pvstereo,huang2022h,godard2017unsupervised} tasks. Nevertheless, there has been a notable shift towards the adoption of ViT-based architectures in Computer Vision. Although some works seek to provide explanations for differences between ViTs and CNNs or the effectiveness of ViT techniques~\citep{park2022vision,xie2023revealing}, many ViTs proposals compare their architectures with basic CNNs baselines, such as the ResNet, as recently reported in~\citep{liu2022convnet,smith2023convnets}. Furthermore, certain ViT studies fail to include the runtime of the proposed methods when comparing them to CNN strategies and fail to address the trade-off between metric performance and runtime.

\rev{This trade-off is particularly evident in the field of self-supervised stereo depth, where prior methods typically face two fundamental limitations. 
First, \emph{computational burden}: Transformer-based approaches that attend across entire epipolar lines (e.g., ChiTransformer~\citep{su2022chitransformer}) achieve state-of-the-art metrics but at the expense of runtime and memory, restricting their use in real-time or embedded applications. 
Second, \emph{metric performance}: lightweight CNN-based designs (e.g., Monodepth2~\citep{godard2019digging} and ES$^3$Net~\citep{fang2023es3net}) offer good latency but tend to underperform in accuracy, especially in low-texture or occluded regions. 
Thus, the field continues to face the challenge of achieving both efficiency and competitive accuracy.}

To address the need to achieve state-of-the-art metrics while not increasing the model execution time, we introduce the \textbf{C}ross-attentional \textbf{C}onvolutional Conv\textbf{NeXt} (CCNeXt) architecture, a novel CNN-based architecture for self-supervised stereo depth estimation.\rev{CCNeXt is explicitly designed to balance metric performance and computational efficiency, surpassing both recent CNN and ViT architectures on KITTI and DrivingStereo (\Cref{fig:1}) while running significantly faster than the state-of-the-art ChiTransformer~\citep{su2022chitransformer}.}

For the encoder, we changed the standard ResNet-based encoder~\citep{godard2019digging,huang2022h} to the smaller version of ConvNeXt~\citep{liu2022convnet} feature extractor with the addition of a bottleneck block in the first convolutional block (stem cell), allowing a better feature representation in the larger skip connection dimension. Our model is trained using a shared-weight network between left-right image pairs, coupled with a novel windowed epipolar cross-attention mechanism, which allows the flow of feature representations between right-left feature pairs. We remodel the dual view training for the decoder to reduce execution time and propose the Individual Contextual Expansive Path (ICEP) to enhance the higher dimension outputs. Our model achieves state-of-the-art results in the KITTI dataset~\citep{Geiger2012CVPR} for Absolute Relative difference (AbsRel), Squared Relative difference (SqRel), and accuracy with a maximum relative error of 25\% ($\delta < 1.25$) depth metrics compared to the state-of-the-art method ChiTransformer while being more than ten times faster, and achieves state-of-the-art results for all depth metrics compared to monocular, monocular-stereo, and stereo self-supervised techniques on the KITTI improved Ground Truth (IGT) dataset~\citep{uhrig2017sparsity}, and DrivingStereo datasets~\citep{yang2019drivingstereo}. In addition, we conduct a statistical analysis of the results, extending the customary practice of only comparing the average metric results of different models.

\rev{In summary, unlike prior self-supervised stereo methods that either sacrifice runtime for accuracy or vice versa, CCNeXt introduces a modern ConvNeXt backbone, windowed epipolar cross-attention, and a lightweight decoder that together achieve state-of-the-art accuracy while reducing execution time by an order of magnitude.}

The main contributions of our work are:
\begin{itemize}
\item We introduce a novel encoder-decoder architecture based on ConvNeXt for self-supervised stereo depth estimation, enhancing feature extraction capabilities through the incorporation of a windowed epipolar cross-attention and a composed stem cell for the encoder, and implemented a novel strategy called ICEP for the decoder. This approach enhances low and high-resolution output quality and metrics, as demonstrated by ablation studies.
\item Our approach achieves state-of-the-art results for three out of seven depth estimation metrics on KITTI while being 10.18$\times$ faster than the current best model and achieves state-of-the-art results for all metrics on the KITTI IGT and DrivingStereo datasets.
\item We performed a statistical analysis to compare the outcomes of our model with the literature results. We also compared published solutions with available training procedures on the DrivingStereo dataset, enabling us to discern the strengths and limitations of the KITTI dataset for depth estimation. 
\end{itemize}

\begin{figure*}[!htb]
  \centering
  \includegraphics[width=0.99\linewidth]
  {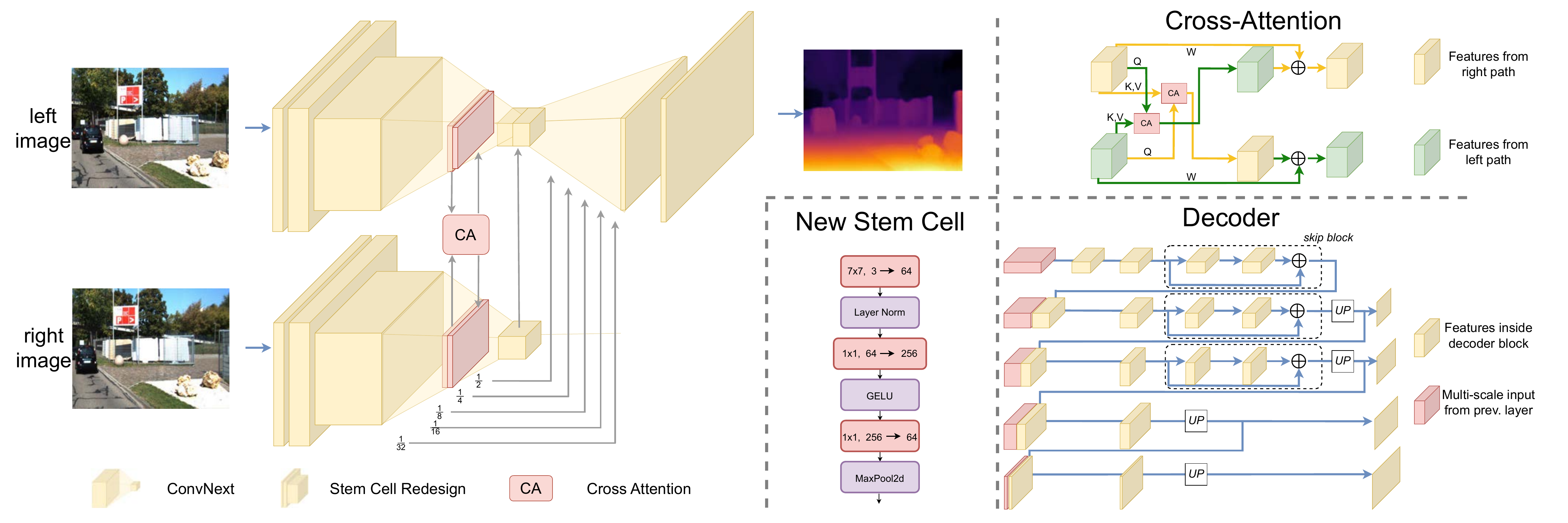}
  \caption{\textbf{Overview of our framework.} We proposed a shared-weight architecture with skip connections and windowed epipolar cross-attention on the Encoder (Left). Details on cross-attention (Top right), Decoder architectures (Bottom right)\rev{, and new stem cell proposal (Center bottom)} are also displayed.} 
  \label{fig:2}
\end{figure*}

\section{Related Work}
\label{sec:related}

Depth estimation methods can be classified based on the number of images used (monocular, stereo, or multi-view) and the type of supervision (supervised or self-supervised). While these approaches share a common goal, they often achieve remarkable results under specific configurations, such as supervised monocular or self-supervised stereo depth estimation.

\rev{This distinction is particularly clear in monocular methods, where supervised and self-supervised techniques diverge significantly.} For instance, Fu \etal~\citep{fu2018deep} proposed a method using ordinal regression to estimate continuous depth through linear combinations of outputs, which was further refined with adaptive and iterative combinations~\citep{shao2024iebins,li2024binsformer}. In contrast, self-supervised techniques typically lack similarities in loss functions, architectures, or strategies with supervised methods. These techniques often estimate depth and camera motion between consecutive frames. Zhou \etal~\citep{zhou2017unsupervised} proposed a pipeline that predicts depth and pose using dedicated networks for each task. Subsequent improvements included multi-scale sampling and auto-masking loss to improve visual artifacts and object motion~\citep{godard2019digging} and scale-aware training for better generalization~\citep{guizilini20203d}. Recently, Guizilini \etal~\citep{guizilini2022multi} proposed a Transformer-based approach that uses a cross-attention matching mechanism instead of predicting camera pose for the input frames.

Supervised and self-supervised stereo vision approaches also differ in loss, architecture, and training strategies. Supervised stereo depth and stereo matching techniques commonly employ a cost volume based on CNN-extracted features from left and right images to create a matching prediction~\citep{xu2023iterative,chang2018pyramid,duggal2019deeppruner,lipson2021raft,xu2022attention, kusupati2020normal}. These are regularized to produce better representations since this simple cost volume is noise-contaminated~\citep{laga2020survey}. There are many techniques to regularize, where some works use Convolutional Gated Recurrent Units (ConvGRUs) to update the disparity maps using local cost values~\citep{xu2023iterative, lipson2021raft}, or using 2D convolutions~\citep{duggal2019deeppruner}. Transformer-based architectures were also proposed~\citep{guo2022context,xu2023unifying,li2021revisiting}, where global context information is extracted using a combination of cross-attention and axial attention~\citep{guo2022context}, and a unified model was proposed for rectified stereo matching, unrectified stereo depth estimation, and optical flow~\citep{xu2023unifying}. \rev{Therefore, these architectures are typically designed to enhance fine-grained prediction details under a supervised loss function, often by incorporating complex decoders with various cost volume compositions. However, as we demonstrate in Section~\ref{sec:kitti}, they tend to yield poor representations in self-supervised settings.}

\rev{For self-supervised stereo methods,~\citet{godard2017unsupervised} introduced} left-right consistency, generating depth maps for both left and right images using a combination of consistency loss, disparity smoothness loss, and photometric reprojection error loss. Besides being applied for monocular depth estimation, Monodepth2~\citep{godard2019digging} also presented setups for self-supervised stereo depth estimation and a monocular-stereo setup, mixing left-right information with video frames.

More recently, H-Net~\citep{huang2022h} has proposed finding epipolar features by matching candidates using optimal transport as part of an attention mechanism during the encoder and decoder stages, allowing features from the left and right paths to be shared in a siamese encoder-decoder network.~\citet{su2022chitransformer} proposed ChiTransformer, a Transformer-based approach that applies multiple self-attention and cross-attention layers for left-view and right-view features extracted using a CNN and fuses information back in a decoder using convolutions. ChiTransformer presents different strategies, including a gated positional cross-attention mechanism that limits the attentional area to match the epipolar geometry. Recently, ES$^3$Net~\citep{fang2023es3net} used the RealtimeStereo~\citep{chang2020attention} as a backbone architecture, adapting it to a self-supervised training using two left-right flipped inputs to predict left and right disparity maps. \rev{Together, these approaches illustrate the broader trade-off discussed in the Introduction: CNN-based designs are generally efficient but lag behind in accuracy, whereas Transformer-based methods achieve strong performance at the expense of runtime and memory.}

ChiTransformer currently holds state-of-the-art performance for self-supervised depth estimation across various metrics. However, its runtime is compromised by the presence of multiple attention modules. \rev{To overcome this bottleneck, we propose CCNeXt, which integrates a modern CNN backbone with a novel windowed epipolar cross-attention mechanism and a lightweight decoder. This design achieves state-of-the-art accuracy while significantly reducing execution time.}

\rev{Beyond stereo depth, self-supervised and weakly supervised learning have progressed rapidly across domains beyond stereo depth. Examples include federated contrastive learning with feature distillation for human activity recognition~\citep{FCL-HAR-2021} and contrastive token and label activation for remote-sensing semantic segmentation~\citep{CTLA-RS-2022}. 
While these works share our overarching objective of learning strong representations with limited supervision, CCNeXt differs in two fundamental ways. 
First, our supervision signal is geometry-driven: stereo photometric reprojection under known intrinsics and baseline, in contrast to instance- or feature-level contrastive objectives and augmentation strategies without geometric grounding. 
Second, our architecture explicitly encodes epipolar awareness through windowed cross-attention constrained by a physically valid disparity range, which is unique to stereo geometry. 
Thus, CCNeXt is complementary: methods such as contrastive pretraining or token selection could be integrated into our encoder, but our core novelty lies in geometry-constrained attention that simultaneously enhances efficiency and matching reliability.}

\section{Methodology}

In this section, we describe our CCNeXt architecture, detailing our encoder composed of the backbone choice, stem cell substitution, windowed cross-attention mechanism, and our proposed decoder architecture with changed convolutional layer positions and skip blocks. The complete architecture is shown in~\Cref{fig:2}.

\subsection{Encoder}

\textbf{ConvNeXt.} We initially upgraded the ResNet encoder backbone architecture used by Monodepth2, ChiTransformer, and H-Net strategies. We chose ConvNeXt for its equivalent performance to Transformer-based architectures such as Swin Transformer~\citep{liu2021swin}. Besides using ResNet-based architectures, Monodepth2 and H-Net use a lighter ResNet18 version than the ResNet50 version adopted by ChiTransformer. Thus, we chose a ConvNeXt version that is slightly faster and has approximately 10\% fewer parameters than ResNet18.


\textbf{Stem Cell.} For our ConvNeXt architecture, we propose the usage of additional convolutional layers, substituting the stem cell of the original ConvNeXt to extract deeper features. This allows our architecture to have more complex features before skip connecting this information to the decoder. The first skip connection is the closest one to the highest feature map dimensions. Therefore, it directly affects the quality of the final prediction. The proposed Stem Cell structure follows the structure of the Inverted Bottleneck that is applied in each block of the ConvNeXt~\citep{liu2022convnet}. \rev{Details of the implementation, including convolutional layer parameters (kernel sizes, input and output channels) and the ordering of additional layers such as normalization and activation, are provided in \Cref{fig:2}.}

\begin{figure}[!htb]
  \centering

  \includegraphics[width=1.0\linewidth]{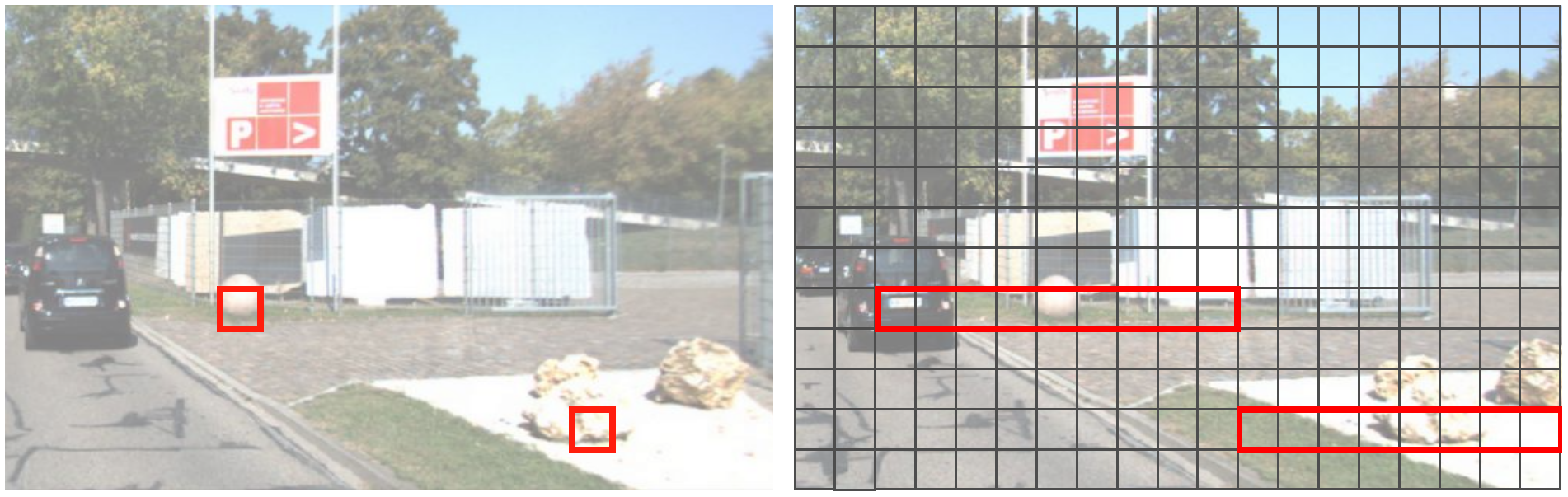}
  \caption{\textbf{Windowed Epipolar Cross-Attention.} We proposed a windowed cross-attention mechanism that limits the attention to valid epipolar candidates. Points of interest in an image (Left) can only be matched with a limited group of pixels $w_{pl}$ on the other stereo image.} 
  \label{fig:wind}

\end{figure}

\textbf{Windowed Cross-Attention.} The proposed windowed cross-attention mechanism (\Cref{fig:wind}) is inspired by~\citet{huang2022h} in its EG-MEA approach of H-Net, MVSTER~\citep{wang2022mvster}, and Swin Transformer~\citep{liu2021swin}. H-Net is inspired by~\citet{wang2018non} self-attention, which adapts the Transformer-based mechanism to CNNs using 1$\times$1 convolutions instead of using the projection matrices for key, query and value. The principal particularity of our proposed cross-attention mechanism is that it is a windowed epipolar attentional mechanism; therefore, only features within a limited range along the same row of the left-view $X_l$ and right-view $X_r$ features are evaluated as matching candidates for Key and Query components. \rev{This restriction greatly reduces the likelihood of false positives between visually similar structures that cannot correspond geometrically. 
For instance, this is particularly beneficial when multiple traffic lights appear at the same height but on opposite sides of a road.}

\rev{
To further constrain the attention, we exploit the stereo geometry to define a maximum search range. 
Given a rectified stereo pair $I^l, I^r \in \mathbb{R}^{H\times W}$ with focal length $f_x$ and baseline $B$, the largest disparity at encoder scale $s$ (downsampling factor $2^s$) can be estimated from a minimum plausible scene depth $D_{\min}$ as in Equation~\ref{eq:disp_bound}.}

\begin{equation}
\Delta_{\max}(s) = \frac{B f_x}{D_{\min} \, 2^s}.
\label{eq:disp_bound}
\end{equation}

\rev{Instead of attending across the entire image row, each pixel only considers candidates within $\pm \Delta_{\max}(s)$ around its location. 
This yields a window of width $w_s = 2\Delta_{\max}(s)+1$, which is typically much smaller than the full row. 
As a result, the attention cost per query is reduced from $O(W/2^s)$ in full-row attention to $O(w_s)$ in our formulation. 
This windowing not only lowers the computational complexity but also prevents spurious correspondences at geometrically implausible disparities (e.g., multiple traffic lights aligned at the same height but far apart in the scene).
}

In Equation~\ref{eq:1}, we define the cross-attentional mechanism, where $Q_r = q(X_r)$, $K_l = k(X_l)$, and $V_l = v(X_l)$ are defined in $Q_r \in \mathbb{R}^{c_{in}\times w \times h}$, $K_l \in \mathbb{R}^{c_{in}\times w \times h}$, and $V_l \in \mathbb{R}^{c_{in}\times w \times h}$. The $q$, $k$, and $v$ are 1$\times$1 convolutions that map input of dimension $\left (c \times h \times w\right )$ to $\left (c_{in}\times h\times w\right )$.
\begin{equation}
  \text{Att}_l(Q_r,K_l,V_l) = \text{softmax}(Q_rK_l^T) V_l
  \label{eq:1}
\end{equation}

The $\text{Att}_l$ softmax term maps every $w^r \times w_{pl}^l$, making it possible to find the correspondent $w_{i}^r$ position for each $w_{i}^l$ element in each windowed row of the feature map. Subsequently, we apply an additional convolutional layer $W_l$ as the last linear layer of the original self-attention mechanism. This output is combined with the right-view input features $X_r$ as available in Equation~\ref{eq:2}.
\begin{equation}
Y_r = W_l\left (\text{Att}_l \right ) + X_r
  \label{eq:2}
\end{equation}

We did not use multi-head attention due to its computational cost increase, so our attention can be interpreted as the multi-head attention layer proposed by~\citet{vaswani2017attention} with the number of heads equal to one. This architecture is also mirrored for the attentional outputs of $Y_l$, where now we have $Q_l$, $K_r$, and $V_r$, instead of $Q_r$, $K_l$, and $V_l$.

\subsection{Decoder}
\label{dec}

For the decoder, we introduced the ICEP architecture (\Cref{fig:2}) with added convolutional blocks with skip connections instead of having cross-attention modules as performed in H-Net. This modification enables our architecture to predict a single-view disparity estimation using a pair of inputs, as there is no interdependence between the right and left depth predictions in the decoder. Consequently, this design choice results in a significant reduction of execution time and memory usage, nearly halving both. Similarly to H-Net, before starting the decoder stage, we concatenate features from both left and right paths to form a ($2C \times H \times W$) volume that is then reduced to a ($C \times H \times W$) feature map by applying a 3$\times$3 convolutional layer. This volume is similar to the cost volume used in the stereo matching approaches. However, we do not concatenate different dimension features ($D_n$) to form a single ($2 C D_n \times H \times W$) feature map. Consequently, we form $D_n$ features of ($C_i \times H_i \times W_i$), where $i$ varies depending on the input feature map size. 

\textbf{Intra Skip Blocks.} We propose the addition of convolutional blocks, each with their own skip connections, within the decoder path (\Cref{fig:2}). This strategy is intended to enhance the representation, preserve the backpropagation of gradient in low-resolution features, and break the left-right decoders' dependency during inference, which reduces the execution time of the model. The skip blocks apply two convolutions with the same channels as the previous layers in each $i$-th layer. We did not use the intra skip blocks in the higher resolution layers of the decoder since it would increase the computational cost of our decoder. In the lower resolution blocks, this increment is derisory.

\textbf{ICEP.} \rev{The final ICEP decoder modifies the Monodepth-style architecture~\citep{godard2017unsupervised} with three key changes: 
(i) a symmetric high-resolution head that prevents the highest-scale features from supervising multiple output scales simultaneously; 
(ii) delayed fusion of the top-level skip ($x_0$), so that $y_0$ alone benefits from the finest encoder details, improving boundary sharpness; 
(iii) intra-decoder skip blocks (two $3{\times}3$ convolutions with residual connection) at lower-resolution stages, which preserve gradient flow and suppress phantom responses in textureless regions. 
These design changes allow for more accurate depth prediction across all scales.
}

\begin{figure}[!htb]
\centering
\begin{minipage}{\columnwidth}
\resizebox{0.86\textwidth}{!}{

\begin{tikzpicture}[x=0.75pt,y=0.75pt,yscale=-1,xscale=1]

\draw (181.08,275.53) node  {\includegraphics[width=222.99pt,height=67.81pt]{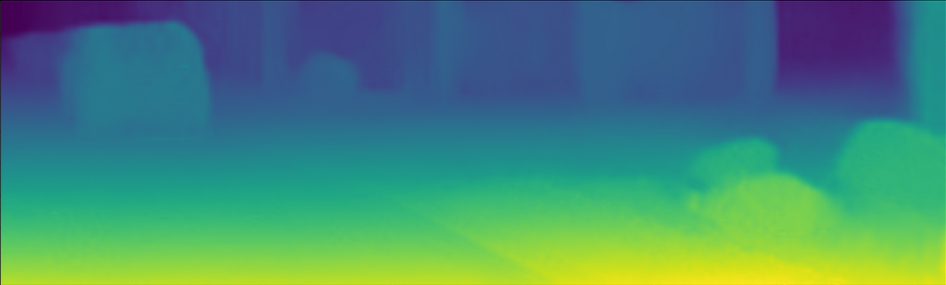}};
\draw (181.08,169.5) node  {\includegraphics[width=222.99pt,height=67.25pt]{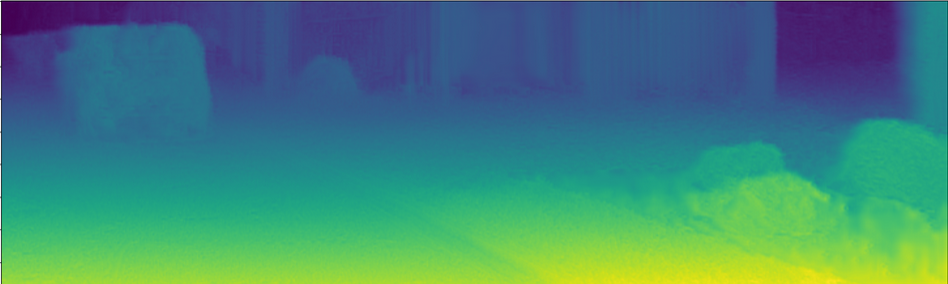}};
\draw (181.08,62.86) node  {\includegraphics[width=222.99pt,height=67.81pt]{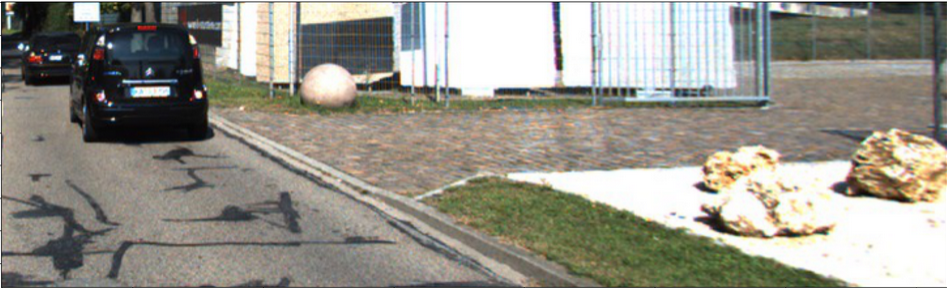}};
\draw   (32.42,17.65) -- (329.74,17.65) -- (329.74,108.07) -- (32.42,108.07) -- cycle ;
\draw   (32.42,230.32) -- (329.74,230.32) -- (329.74,320.74) -- (32.42,320.74) -- cycle ;
\draw   (32.42,124.32) -- (329.74,124.32) -- (329.74,214.74) -- (32.42,214.74) -- cycle ;
\draw  [color={rgb, 255:red, 255; green, 0; blue, 0 }  ,draw opacity=1 ][line width=1.5]  (32.42,230.32) -- (144.33,230.32) -- (144.33,289.67) -- (32.42,289.67) -- cycle ;
\draw  [color={rgb, 255:red, 255; green, 0; blue, 0 }  ,draw opacity=1 ][line width=1.5]  (225,251) -- (329.74,251) -- (329.74,320.74) -- (225,320.74) -- cycle ;
\draw  [color={rgb, 255:red, 255; green, 0; blue, 0 }  ,draw opacity=1 ][line width=1.5]  (225,144.59) -- (329.74,144.59) -- (329.74,214.33) -- (225,214.33) -- cycle ;
\draw  [color={rgb, 255:red, 255; green, 0; blue, 0 }  ,draw opacity=1 ][line width=1.5]  (32.42,124.32) -- (144.33,124.32) -- (144.33,183.67) -- (32.42,183.67) -- cycle ;
\draw  [color={rgb, 255:red, 255; green, 0; blue, 0 }  ,draw opacity=1 ][line width=1.5]  (225,38.33) -- (329.74,38.33) -- (329.74,108.07) -- (225,108.07) -- cycle ;
\draw  [color={rgb, 255:red, 255; green, 0; blue, 0 }  ,draw opacity=1 ][line width=1.5]  (32.42,17.65) -- (144.33,17.65) -- (144.33,77) -- (32.42,77) -- cycle ;

\draw (5.68,79.85) node [anchor=north west][inner sep=0.75pt]  [rotate=-270] [align=left] {Input};
\draw (5.68,308.03) node [anchor=north west][inner sep=0.75pt]  [rotate=-270] [align=left] {ICEP+Skip};
\draw (5.68,220.53) node [anchor=north west][inner sep=0.75pt]  [rotate=-270] [align=left] {From Monodepth2};

\end{tikzpicture}

}
\end{minipage}
  \caption{On \textbf{top}: Input image from the KITTI dataset. On \textbf{middle}: Disparity estimation using the same encoder and the decoder of Monodepth2. On \textbf{bottom:} Same encoder as in the middle with the ICEP decoder with Skip blocks.}
  
  \label{fig:5}

\end{figure}
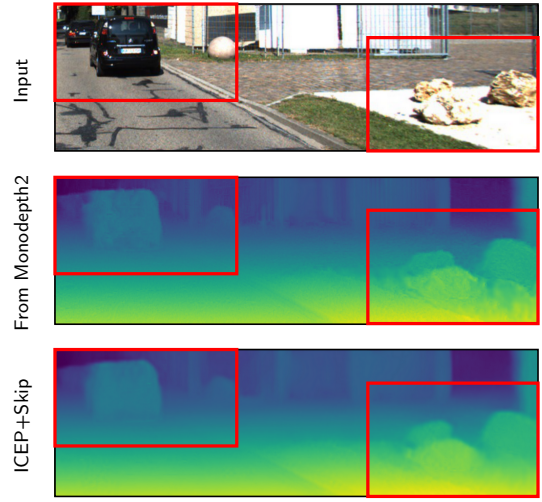

\rev{In Monodepth, skip connection concatenations start from the beginning of the decoder, meaning the last two blocks share almost the same features except for an upsampling and extra convolutions. As a result, high-resolution features ($x_0$) influence both $y_1$ and $y_0$, which disturbs fine-grained detail learning. In ICEP, we postpone the $x_0$ fusion so that only $y_0$ uses the finest-scale features, producing consistent and more detailed outputs (see \Cref{fig:5}).}

\rev{For upsampling, we adopt nearest-neighbor interpolation followed by a $3{\times}3$ convolution to avoid checkerboard artifacts, and concatenate encoder skips with decoder activations at each scale before the intra-skip block. This design preserves high-frequency details while keeping computation modest.}



\subsection{Training Procedures}

Both left and right views are trained, producing $\text{disp}_r$ and $\text{disp}_l$ outputs for each view. Similar to~\citep{zhong2017self,huang2022h,su2022chitransformer,godard2017unsupervised,godard2019digging,wang2021pvstereo}, we formulate the problem as a photometric reprojection error, adding an edge-aware smoothness loss, as proposed by Monodepth~\citep{godard2017unsupervised}. Although we predict the disparities as the output of our network, we do not warp the disparity information directly on the input image $I_i$ to generate the other view image $I^{\text{rep}}_j$. We first convert our disparity map into a dense depth map, and then we perform the camera projection from one view to the other using the intrinsic parameters and extrinsic parameters of the camera system. For the extrinsic parameters, we only add the translation caused by the distance of the cameras in the stereo system, which is the baseline distance. The same procedure is performed in most self-supervised depth estimation or stereo matching techniques~\citep{wang2021pvstereo,huang2022h,godard2019digging,su2022chitransformer}.

The photometric reprojection error ($\mathcal{L}_{\text{pe}}$) is formed by the combination of the SSIM metric~\citep{wang2004image} and the L1 loss, as available in Equation~\ref{eq:3}, where $\alpha=0.85$. We also apply the auto-masking strategy proposed by Godard \etal~\citep{godard2017unsupervised}, filtering out pixels where the change between left and right original views ($I^l$ and $I^r$) is smaller than the difference of the predicted reprojection and the original image ($I^l$ and $I^l_{\text{pred}}$). 
\begin{equation}
\label{eq:3}
\begin{split}
\mathcal{L}_{\text{pe}}\left (I^l, I^l_{\text{proj}}\right) = \frac{\alpha}{2} \left (1 - \text{SSIM}(I^l,I^l_{\text{proj}}) \right ) +\\
(1-\alpha) \left \| I^l - I^l_{\text{proj}} \right \|
\end{split}
\end{equation}

The edge-aware smoothness loss term is composed of the mean-normalized disparity $d^*_t = d_t / \overline{d_t}$, and is available in Equation~\ref{eq:4}. 
\begin{equation}
\label{eq:4}
\begin{split}
\mathcal{L}_{\text{sm}}\left (I^{\text{proj}}, d^* \right ) = \frac{1}{N} \sum_{i,j} \left | \partial_x d_{ij}^* \right | e^{- \left \| \partial_x I_{ij}^{\text{proj}} \right \|} + \\
\left | \partial_y d_{ij}^* \right | e^{- \left \| \partial_y I_{ij}^{\text{proj}} \right \|}
\end{split}
\end{equation}

The final loss function comprises the right and left smoothness and reprojection errors, as in Equation~\ref{eq:5}, where $m$ is the total number of scales. Although having both depths, we did not use left-right disparity consistency loss since it did not improve results in any evaluated scenario. \rev{For the smoothness loss term, we adopt the same weighting factor $\gamma = 1 \times 10^{-3}$ as reported in Monodepth2. We also experimented with values an order of magnitude lower and higher, but observed no improvement across any evaluated scenario.}
\begin{equation}
\label{eq:5}
\begin{split}
    \mathcal{L}_{\text{total}} = \frac{1}{2m} \sum_{i=0}^{m-1} \mathcal{L}_{\text{pe}}\left (I^l, I^l_{\text{proj}} \right) + \gamma \mathcal{L}_{\text{sm}}\left (I^l_{\text{proj}}, d^*_l \right) + \\
    \mathcal{L}_{\text{pe}}\left (I^r, I^r_{\text{proj}} \right) + \gamma \mathcal{L}_{\text{sm}}\left (I^r_{\text{proj}}, d^*_r \right)
\end{split}
\end{equation}

The relationship of depth estimation and stereo matching in a rectified pair of images is defined as $\text{depth} = Bf_x /{\text{disp}}$, where $B$ is the baseline distance between the cameras (0.54 meters for both used datasets) and $f_x$ is the focal length of the camera system in the $x$ direction.

Therefore, using a known camera system, we can define a minimum and maximum depth value and determine a minimum and maximum disparity value that our system will generate.

\section{Experiments}

This section describes the experiments we conducted in our work, emphasizing implementation details and datasets used. In addition to metric average comparison, we perform a statistical analysis of the experiments since the metric differences are shrinking, making it impractical to compare research papers' metrics solely on the examination of the average of prediction metrics. This may occur due to the extensive usage of a single dataset (KITTI~\citep{Geiger2013IJRR}), its inability to evaluate different weather and illumination conditions~\citep{wang2021pvstereo}, and its low percentage of pixels in distances higher than 40 meters~\citep{yang2019drivingstereo}. Some experiments performed here did not include all the published papers, mainly because weights, models, and training procedures are not available for PFN~\citep{pilzer2019progressive}, Li \etal~\citep{li2021separating}, PVStereo~\citep{wang2021pvstereo}, and SsSMnet~\citep{zhong2017self}. ChiTransformer~\citep{su2022chitransformer} provides both code and weights, but they only present them for a distinct architecture that downsizes the original network, thereby harming the performance metrics when utilizing this shared model. H-Net~\citep{huang2022h} training procedures and weights are also unavailable. However, the authors shared the model and weights over request, enabling it to be statistically compared. Nevertheless, the lack of training procedures does not allow it to be expanded to different datasets or dimensions.

For this reason, we made available the training code, model, and weights of our presented model, allowing the reproducibility of the results presented in this paper.

\subsection{Implementation Details}

Our model was implemented using PyTorch. The model is trained with an input resolution of $1280 \times 384$ pixels for the KITTI experiments and $288 \times 640$ pixels for the DrivingStereo dataset. We performed experiments using batch size 8, 30 epochs, and using the Adam optimizer~\citep{kingma2014adam} for both datasets. We set a learning rate of 1e-4, and after 15 epochs, we decayed its value to 1e-5. We trained our models with four scales, therefore $m=4$ in Equation~\ref{eq:5}. We used the ConvNext `Pico' version for the backbone with a pre-trained version on the ImageNet~\citep{russakovsky2015imagenet} dataset. For a matter of comparison, this architecture presents 8.57M parameters, while the ResNet18 version used in H-Net, Monodepth2, and Monodepth has 11.17M parameters, and the ResNet50 used by ChiTransformer, PFN, and STTR~\citep{li2021revisiting} has 21.28M parameters. The complete CCNeXt architecture has 18.36M parameters. \rev{Our architecture incorporates three window-cross attention layers, inserted after the first three stages of the ConvNeXt encoder to enhance feature interaction across views.}

\begin{table*}[!htb]
\caption{\textbf{Results on KITTI Eigen Split and IGT.} Comparative results with self-supervised methods for KITTI Eigen Split dataset and with different modalities methods for the IGT dataset. In the \textit{Train} column: \textbf{M}: monocular self-supervised, \textbf{MS}: monocular-stereo self-supervised, and \textbf{S}: stereo self-supervised. All methods were trained using the Eigen Split, except for the ES$^3$Net, which used the KITTI Raw. The methods are also classified in terms of input image sizes. Low Resolution (LR) indicates techniques trained using $640 \times 192$, and Full Resolution (FR) ones were trained using dimensions larger than $800 \times 240$. Best results are in \textbf{bold}. \dag Reproducibility codes and weights not available for the IGT. \ddag Shared weights do not reproduce the results from the paper for the KITTI Eigen Split, and the paper did not present results for IGT.}  
\setlength{\tabcolsep}{6pt}
\centering
\label{tab:1}
\resizebox{\textwidth}{!}{%
\begin{tabular}{c c  c  c  c  c  c  c  c  c  c }
\toprule
& \multirow{2}{*}{Method} & \multirow{2}{*}{Train} & \multirow{2}{*}{Input} & \multicolumn{4}{c}{\textit{lower is better}} & \multicolumn{3}{c}{\textit{higher is better}} \\ \cmidrule(lr){5-8} \cmidrule(lr){9-11}

& &  &  & \multicolumn{1}{c}{AbsRel} & \multicolumn{1}{c}{SqRel} & \multicolumn{1}{c}{RMSE} &  \multicolumn{1}{c}{RMSE log} & \multicolumn{1}{c}{$\delta < 1.25$} & \multicolumn{1}{c}{$\delta < 1.25^2$} & \multicolumn{1}{c}{$\delta < 1.25^3$}\\

\hline 
 
\multirow{7}{*}{\rotatebox{90}{KITTI Eigen Split}} &  
MonoDepth2~\citep{godard2019digging} & S  & FR & 0.107 & 0.849 & 4.764 & 0.201 & 0.874 & 0.953 & 0.977 \\

& PFN~\citep{pilzer2019progressive} \dag & S & LR & 0.102 & 0.802 & 4.657 & 0.196 & 0.882 & 0.953 & 0.977 \\

& Li \etal~\citep{li2021separating} \dag & S     & FR & 0.096 & 0.744 & 4.500 & 0.188 & 0.890 & 0.960 & 0.979 \\

& H-Net~\citep{huang2022h} & S  & LR & 0.076 & 0.607 & 4.025 & 0.166 & 0.918 & 0.966 & 0.982 \\

& ChiTransformer~\citep{su2022chitransformer} \dag & S  & FR & 0.073 & 0.634 & \textbf{3.105} & \textbf{0.118} & 0.924 & \textbf{0.989} & \textbf{0.997} \\

& ES$^3$Net~\citep{fang2023es3net} & S  & FR & 0.074 & 0.660 & 4.042 & 0.169 & 0.920 & 0.962 & 0.980 \\

\cline{2-11}

& \textbf{CCNeXt (ours)} & S   & FR & \textbf{0.070} & \textbf{0.581} & 3.883 & 0.171 & \textbf{0.926} & 0.963 & 0.979 \\

\hline
\hline 

\multirow{11}{*}{\rotatebox{90}{IGT~\citep{uhrig2017sparsity}}} & 
 PackNet-SfM~\citep{guizilini20203d} &  M  & FR & 0.071 & 0.359 & 2.723 & 0.092 & 0.959 & 0.992 & 0.998 \\

& DepthFormer~\citep{guizilini2022multi} &  M  & FR & 0.055 & 0.265 & 2.723 & 0.092 & 0.959 & 0.992 & 0.998 \\

& iDisc~\citep{piccinelli2023idisc} &  M  & FR & 0.059 & 0.216 & 2.573 & 0.089 & 0.965 & 0.995 & \textbf{0.999} \\

& DIFFNet+SfM-TTR~\citep{izquierdo2023sfm}  &  M  & FR & 0.056 & 0.273 & 2.600 & 0.093 & 0.969 & 0.992 & 0.997 \\

\cline{2-11}

& Monodepth2~\citep{godard2019digging} &  MS  & FR & 0.080 & 0.466 & 3.681 & 0.127 & 0.926 & 0.985 & 0.995 \\

& SQLdepth~\citep{wang2024sqldepth} &  MS  & FR & 0.052 & 0.223 & 2.550 & 0.084 & 0.971 & 0.995 & 0.998 \\

\cline{2-11}

& Monodepth2~\citep{godard2019digging}  & S & LR & 0.085 & 0.537 & 3.868 & 0.139 & 0.912 & 0.979 & 0.993 \\

& H-Net~\citep{huang2022h} & S & LR &   0.059  &   0.208  &   2.370  &   0.087  &   0.977  &   0.996  &   0.998  \\

& ChiTransformer~\citep{su2022chitransformer} \ddag & S & FR & 0.072  &   0.271  &   2.768  &   0.103  &   0.950  &   0.993  &   0.998   \\

& ES$^3$Net~\citep{fang2023es3net} & S & FR & 0.039  &   0.173  &   2.062  &   0.069  &   0.988  &   0.996  &   0.998  \\

\cline{2-11}

& \textbf{CCNeXt (ours)}  & S  & FR & \textbf{0.031} & \textbf{0.090} & \textbf{1.671} & \textbf{0.055} & \textbf{0.992} & \textbf{0.998} & \textbf{0.999} \\

\hline

\end{tabular}
}
\end{table*}

\subsection{KITTI Dataset}
  
The KITTI dataset~\citep{Geiger2013IJRR} comprises 61 scenes with a total number of 42,406 image pairs available, usually called `KITTI Raw'. Following the literature for monocular and stereo depth estimation~\citep{su2022chitransformer,godard2017unsupervised,godard2019digging,huang2022h,klingner2020self}, we use the procedures introduced by Eigen \etal~\citep{eigen2014depth}, typically called the Eigen Split. This proposal separates 697 images extracted from 29 scenes for evaluation and trains/validates the remaining 32 scenes, which results in 22,600 images for training and 888 for validation. For evaluation, we also used the Garg crop~\citep{garg2016unsupervised}, which mainly removes uninteresting regions, \textit{e.g.}, sky regions. We also employ the KITTI Eigen Split on the Depth Completion dataset~\citep{uhrig2017sparsity}, commonly referred to as the KITTI Improved Ground Truth (IGT) dataset with the Garg crop for evaluation. This official dataset from KITTI includes a semi-dense depth ground truth generated by the authors, with outliers removed from the laser scan in the raw data. We utilized the IGT to evaluate the model, obtaining their ground truth data from the same 697 images as the Eigen Split, whenever available. This yields a testing set comprising 652 images.

All techniques except ES$^3$NET used for comparison in our work were trained with the Eigen Split training set. In contrast, ES$^3$NET was trained on the entire KITTI Raw dataset, which incorporates both the training and testing data from the Eigen Split.

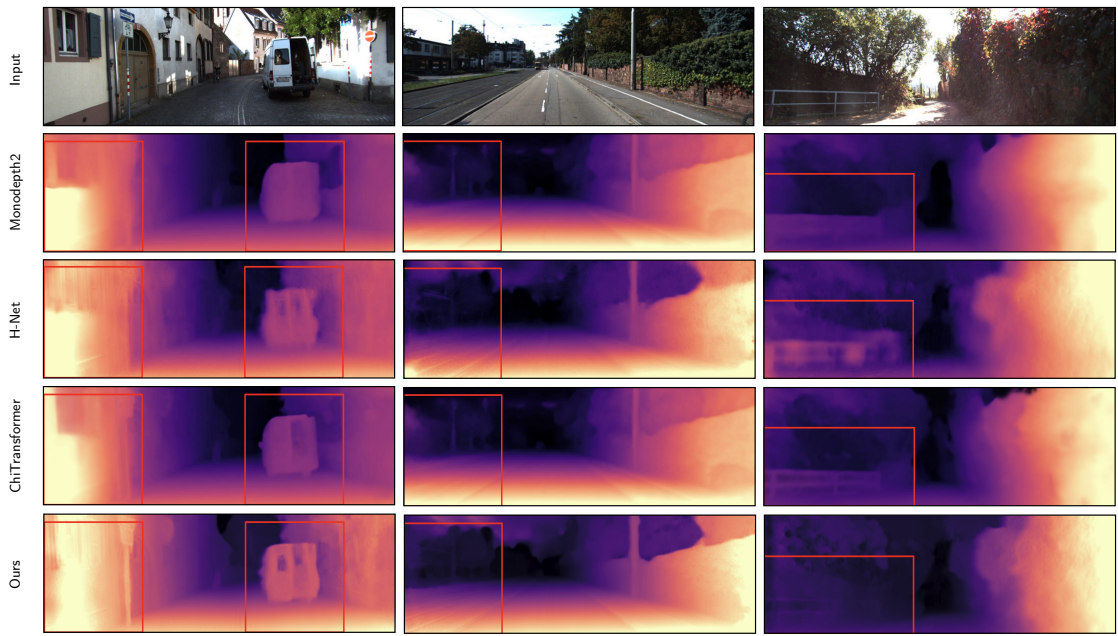
\begin{figure*}[!htb]
  \centering

\resizebox{0.85\textwidth}{!}{
\begin{tikzpicture}[x=0.75pt,y=0.75pt,yscale=-1,xscale=1]

\draw (486.32,486.64) node  {\includegraphics[width=221.2pt,height=74.44pt]{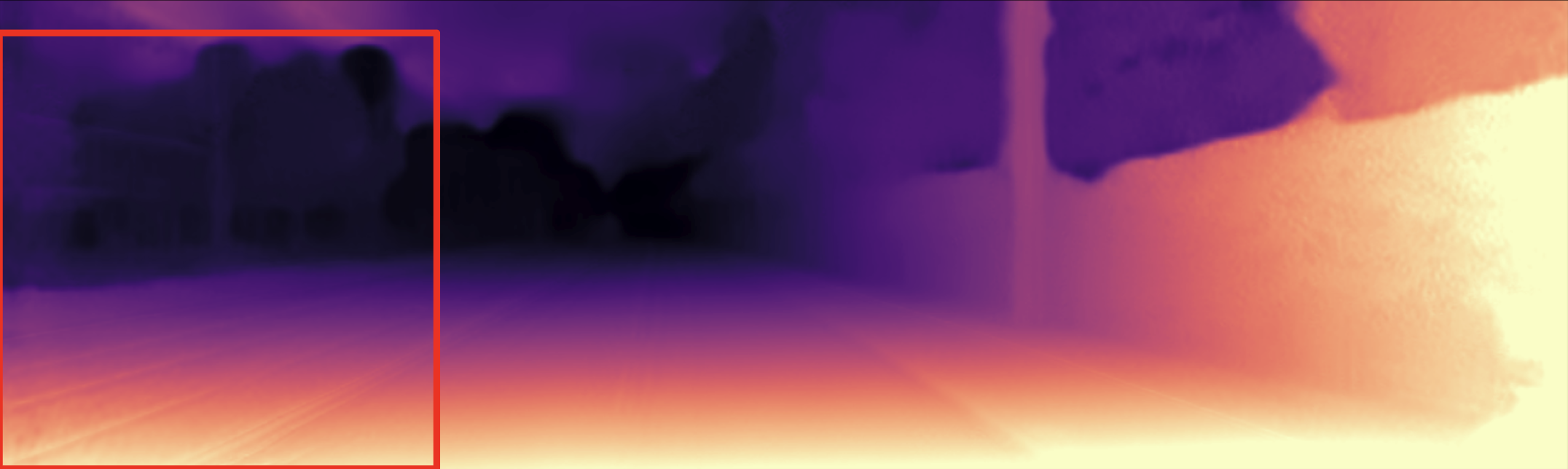}};
\draw (788.69,486.64) node  {\includegraphics[width=221.2pt,height=74.44pt]{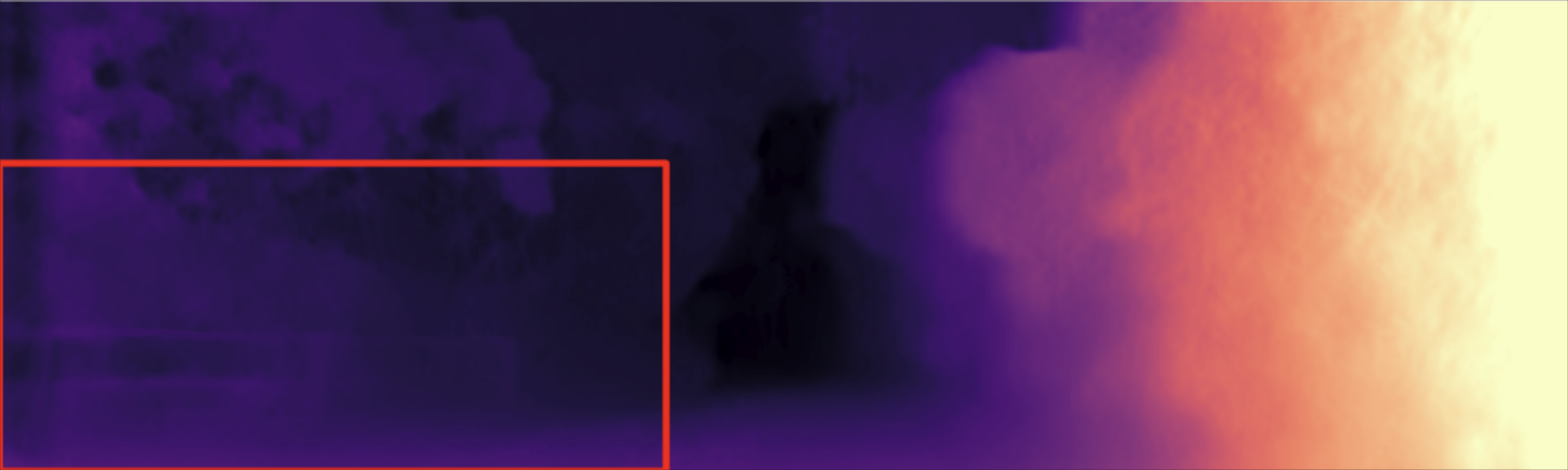}};
\draw (183.5,485.82) node  {\includegraphics[width=221.2pt,height=74.44pt]{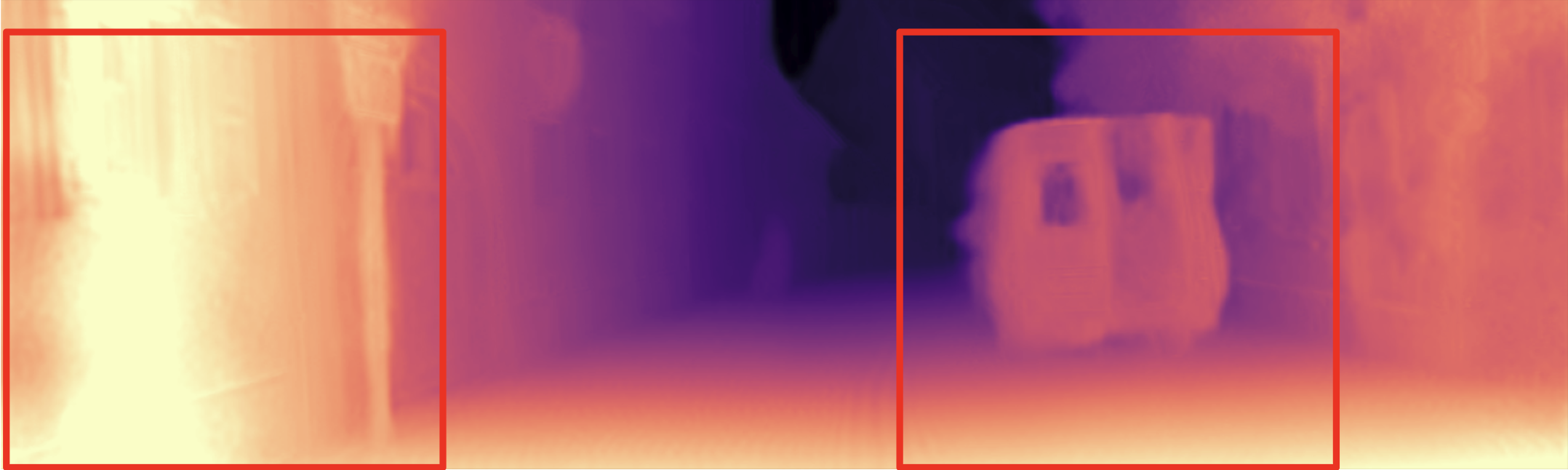}};
\draw (183.15,378.69) node  {\includegraphics[width=221.2pt,height=74.44pt]{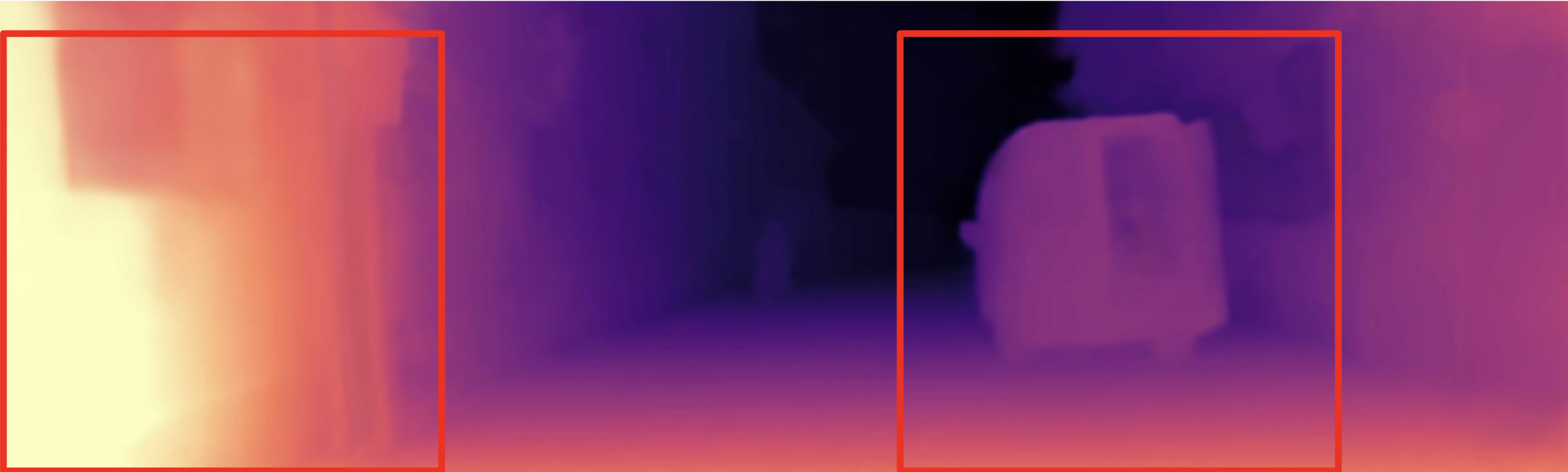}};
\draw (788.35,379.5) node  {\includegraphics[width=221.2pt,height=74.44pt]{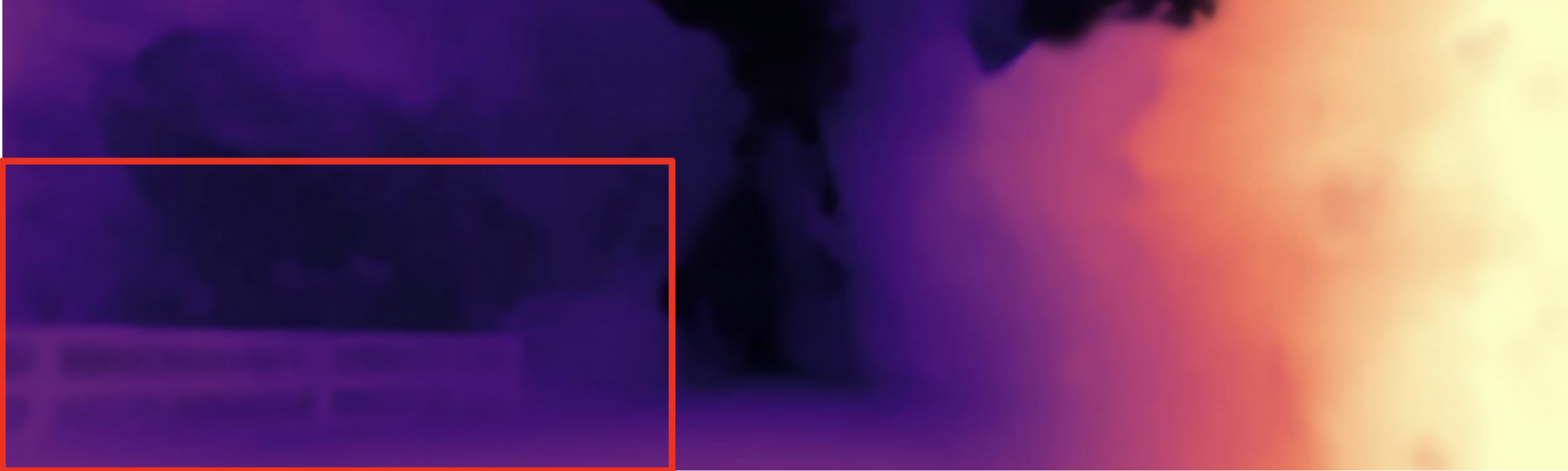}};
\draw (485.97,379.5) node  {\includegraphics[width=221.2pt,height=74.44pt]{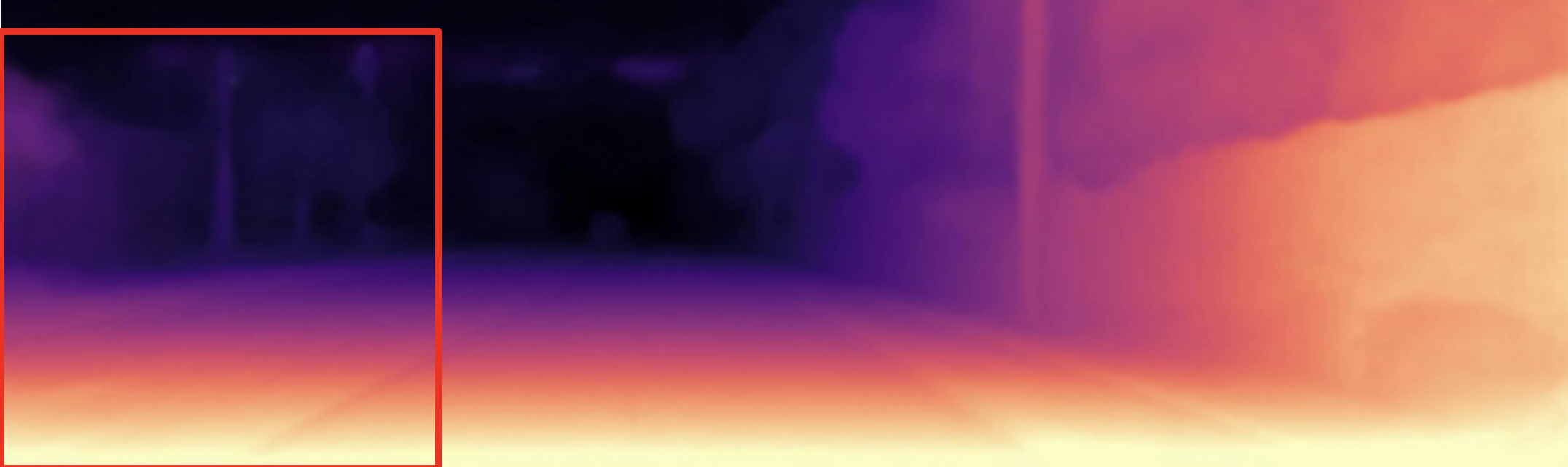}};
\draw (788.35,272.48) node  {\includegraphics[width=221.2pt,height=74.44pt]{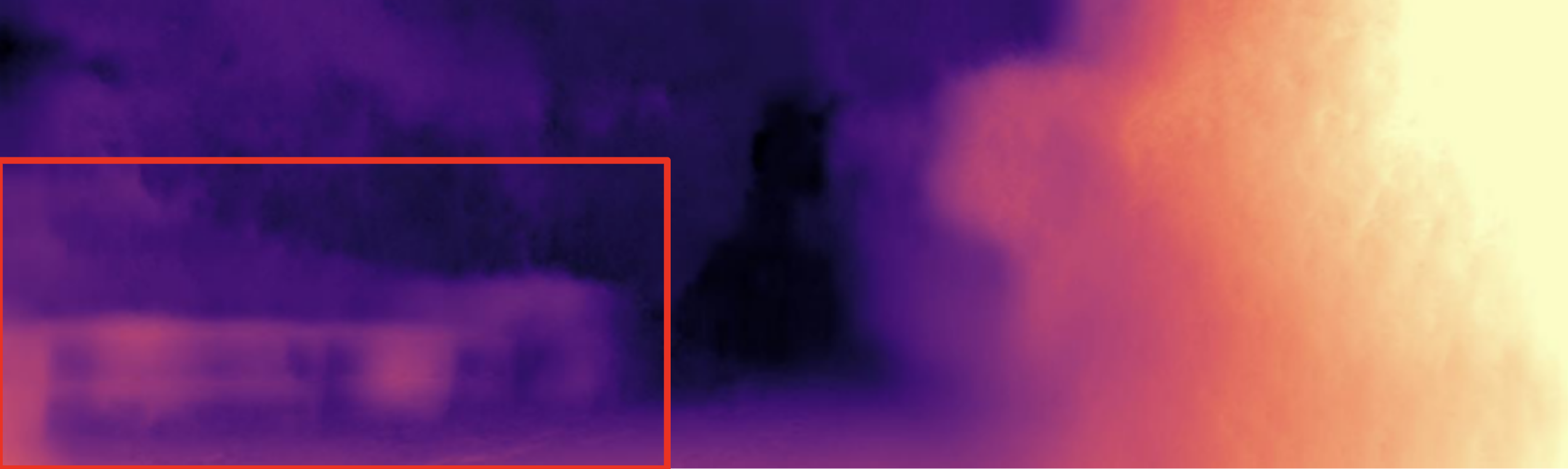}};
\draw (183.15,272.2) node  {\includegraphics[width=221.2pt,height=74.44pt]{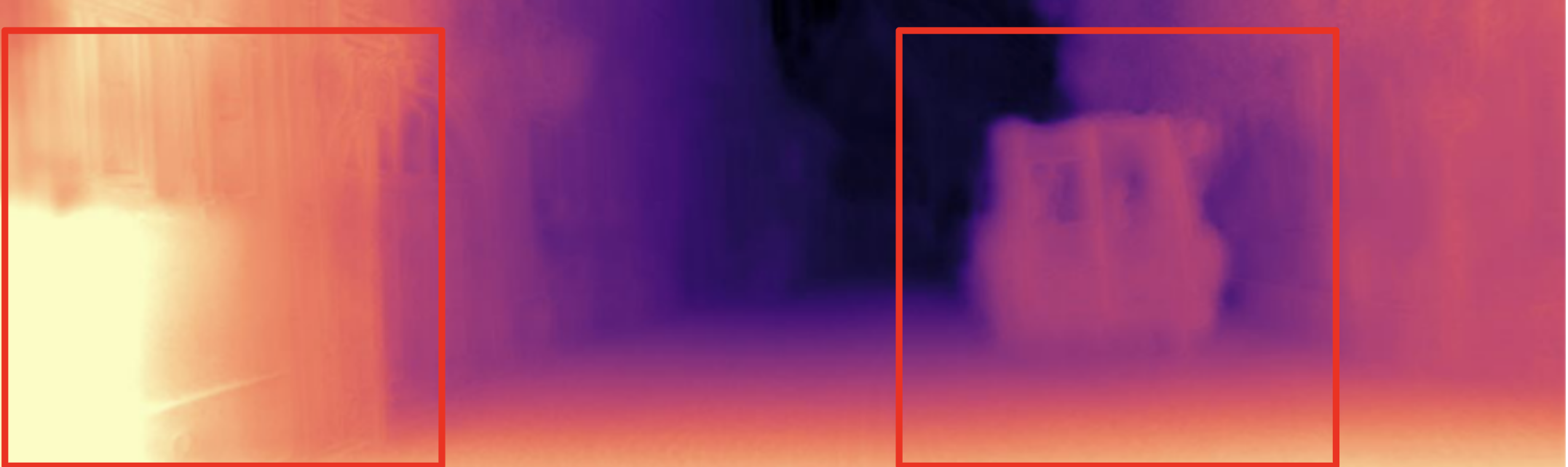}};
\draw (485.97,272.48) node  {\includegraphics[width=221.2pt,height=74.44pt]{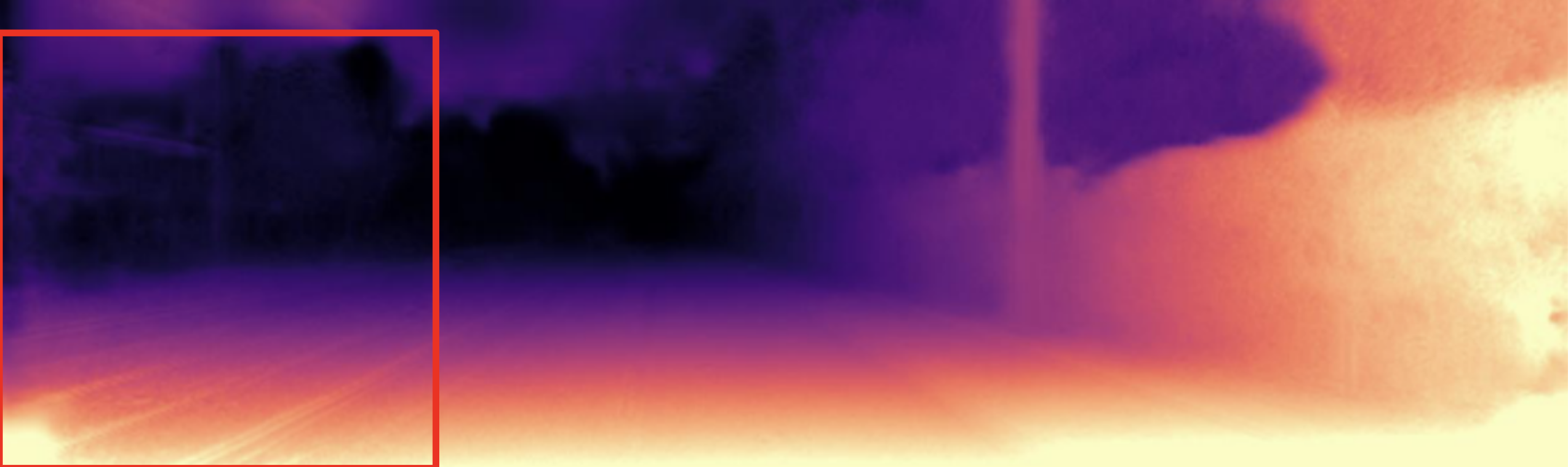}};
\draw (183.15,166.32) node  {\includegraphics[width=221.2pt,height=74.44pt]{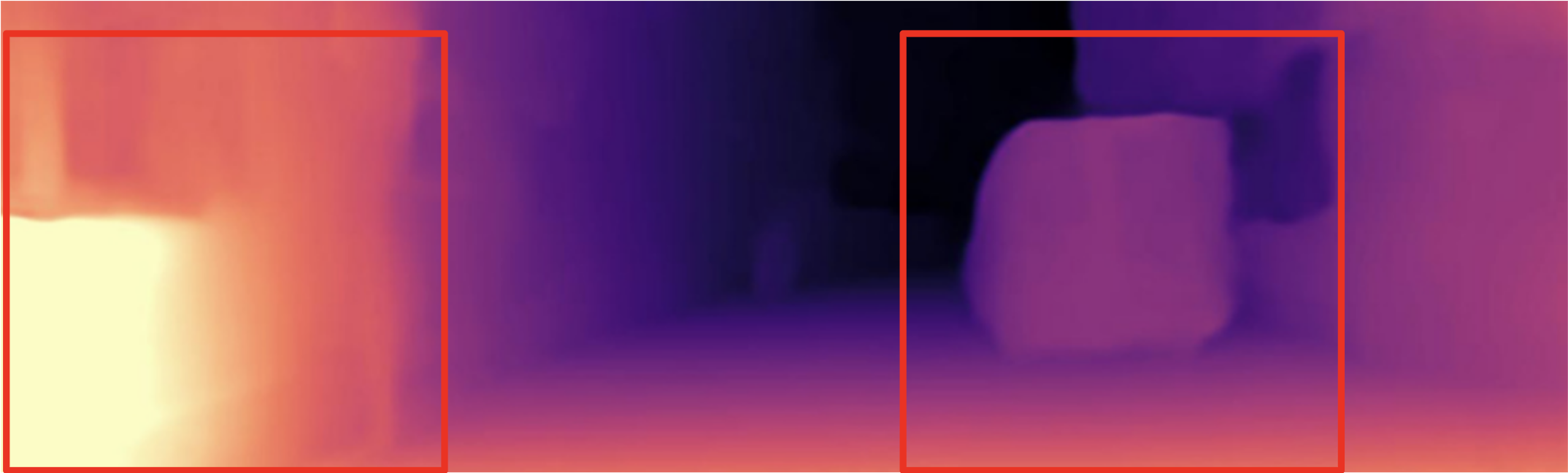}};
\draw (485.28,166.32) node  {\includegraphics[width=221.2pt,height=74.44pt]{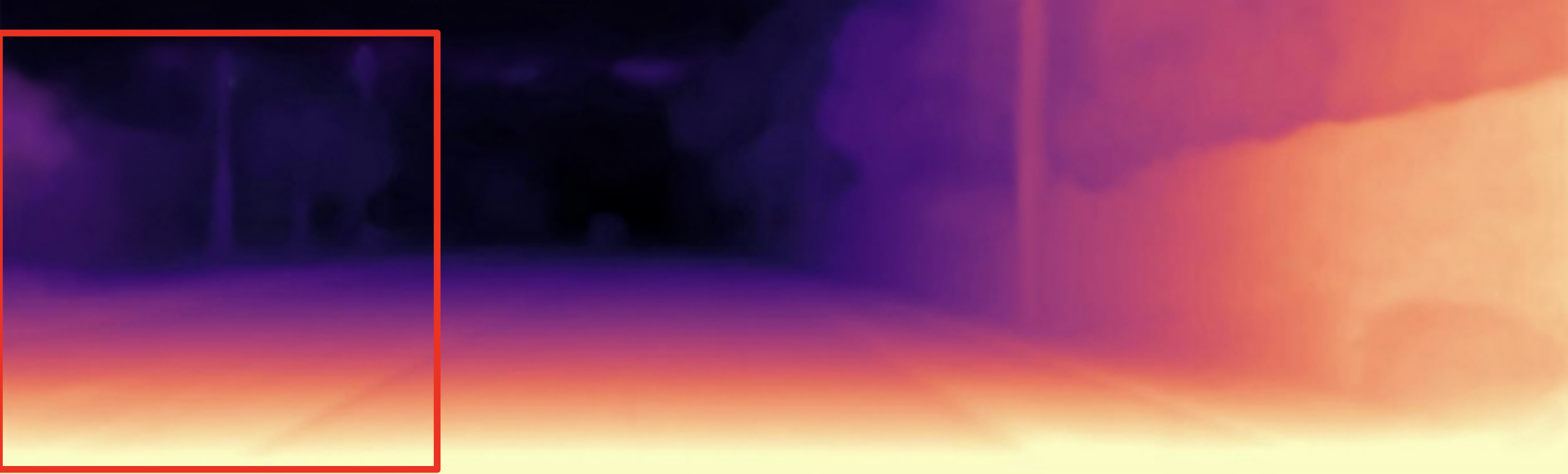}};
\draw (788.73,166.32) node  {\includegraphics[width=221.2pt,height=74.44pt]{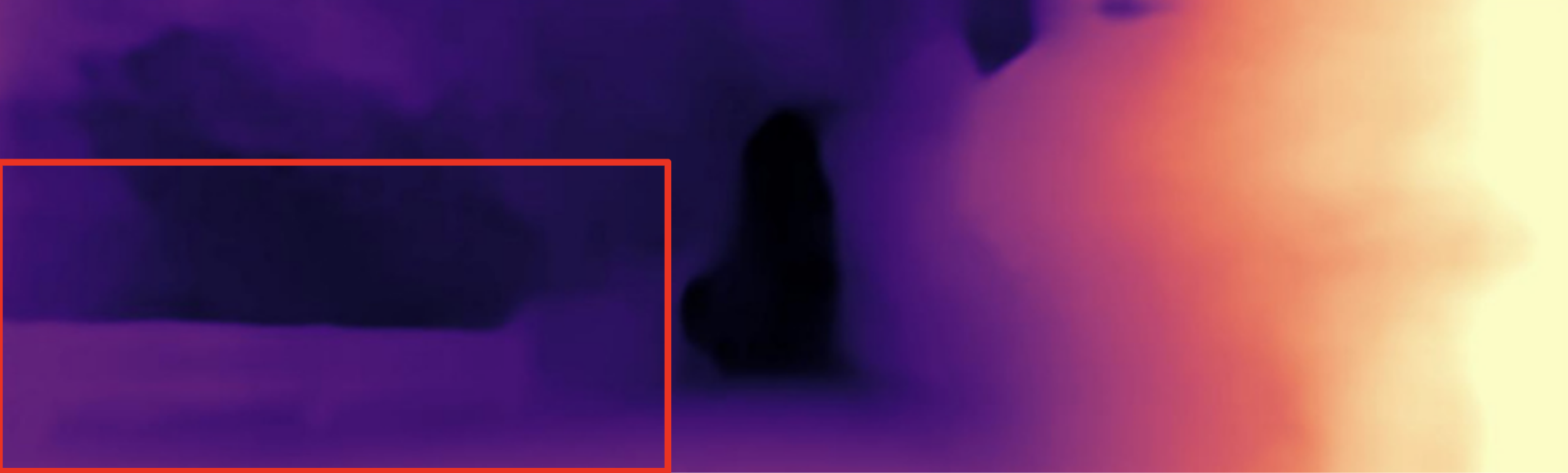}};
\draw (788.3,60.28) node  {\includegraphics[width=221.2pt,height=74.44pt]{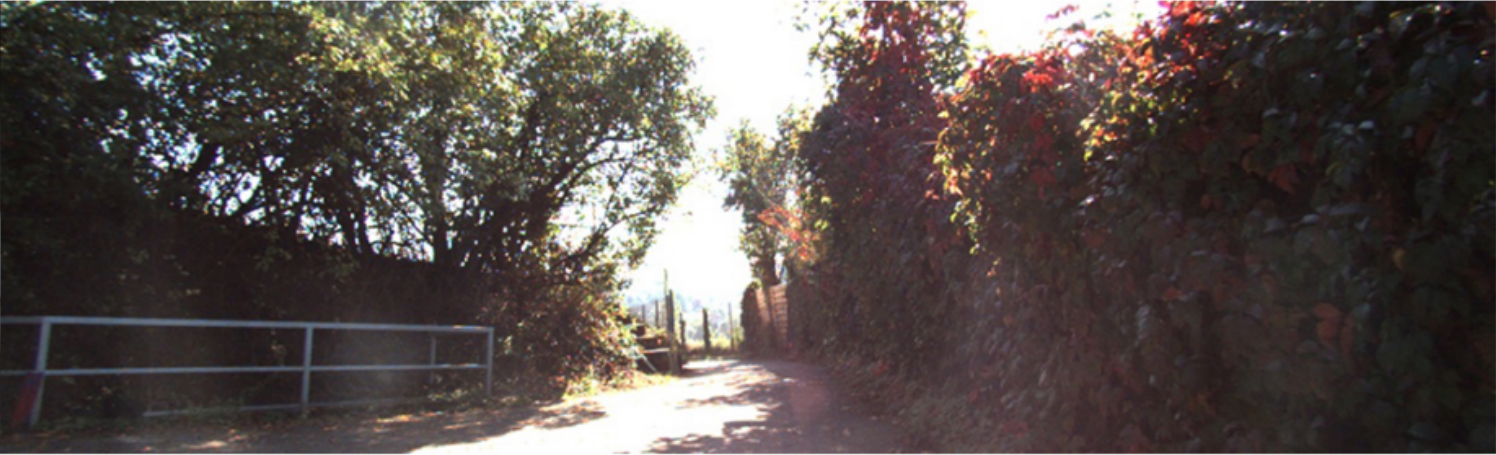}};
\draw (183.15,60.28) node  {\includegraphics[width=221.2pt,height=74.44pt]{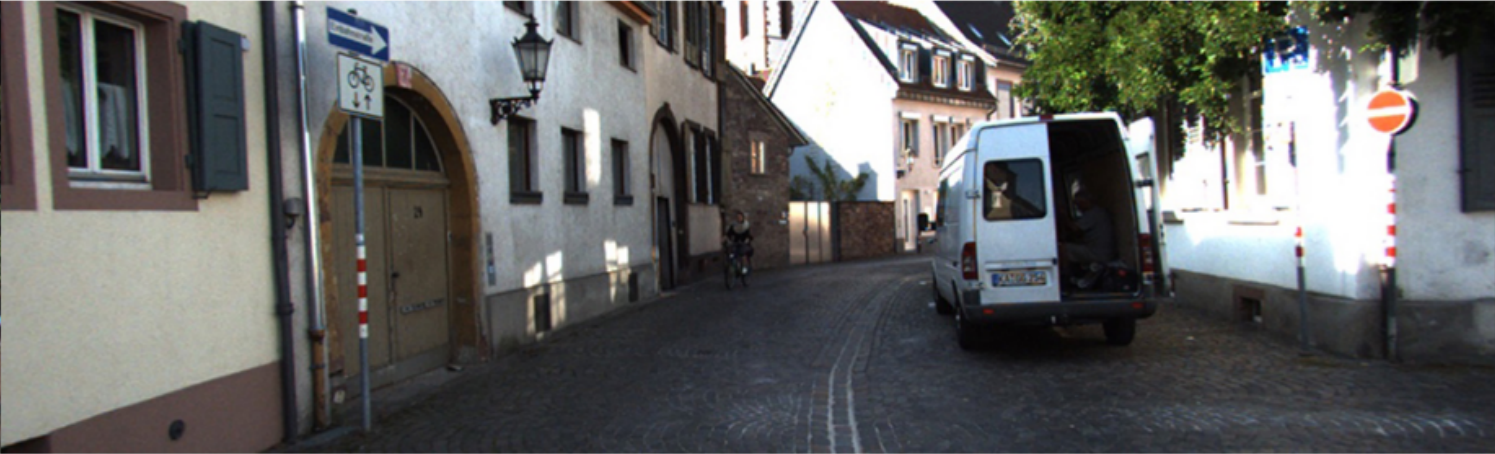}};
\draw (484.85,60.28) node  {\includegraphics[width=221.2pt,height=74.44pt]{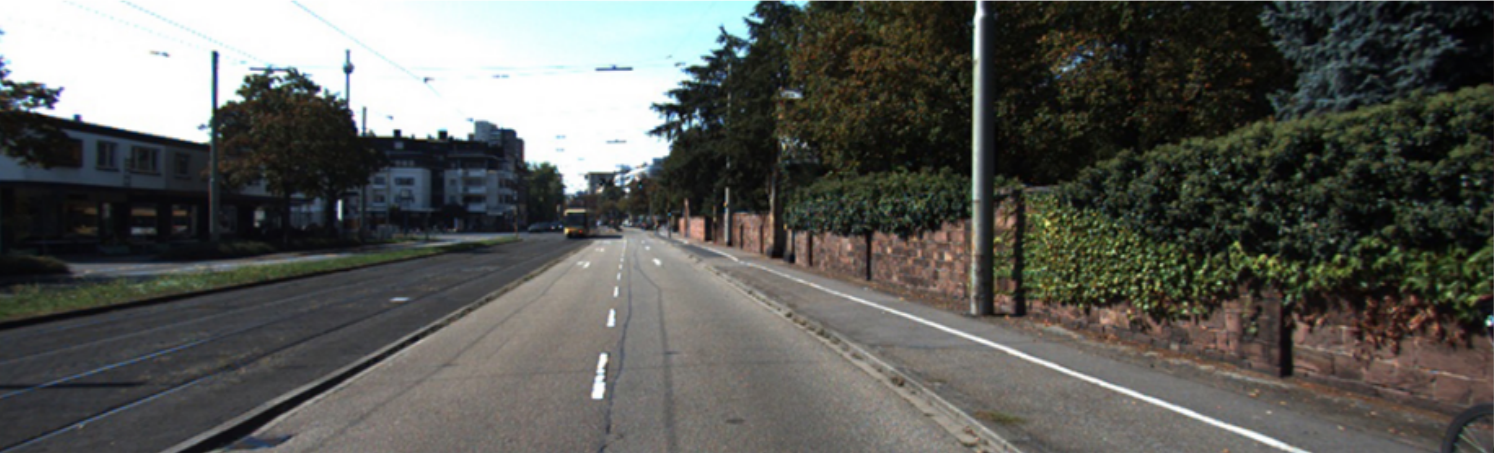}};
\draw   (337.82,116.69) -- (632.75,116.69) -- (632.75,215.95) -- (337.82,215.95) -- cycle ;
\draw   (338.5,222.85) -- (633.44,222.85) -- (633.44,322.1) -- (338.5,322.1) -- cycle ;
\draw   (338.5,329.87) -- (633.44,329.87) -- (633.44,429.13) -- (338.5,429.13) -- cycle ;
\draw   (35.68,329.06) -- (330.62,329.06) -- (330.62,428.31) -- (35.68,428.31) -- cycle ;
\draw   (35.68,222.58) -- (330.62,222.58) -- (330.62,321.83) -- (35.68,321.83) -- cycle ;
\draw   (35.68,116.69) -- (330.62,116.69) -- (330.62,215.95) -- (35.68,215.95) -- cycle ;
\draw   (35.68,10.65) -- (330.62,10.65) -- (330.62,109.91) -- (35.68,109.91) -- cycle ;
\draw   (337.38,10.65) -- (632.32,10.65) -- (632.32,109.91) -- (337.38,109.91) -- cycle ;
\draw   (641.26,116.69) -- (936.2,116.69) -- (936.2,215.95) -- (641.26,215.95) -- cycle ;
\draw   (640.88,222.85) -- (935.81,222.85) -- (935.81,322.1) -- (640.88,322.1) -- cycle ;
\draw   (640.88,329.87) -- (935.81,329.87) -- (935.81,429.13) -- (640.88,429.13) -- cycle ;
\draw   (640.83,10.65) -- (935.77,10.65) -- (935.77,109.91) -- (640.83,109.91) -- cycle ;
\draw   (338.85,437.01) -- (633.78,437.01) -- (633.78,536.26) -- (338.85,536.26) -- cycle ;
\draw   (36.03,436.19) -- (330.96,436.19) -- (330.96,535.45) -- (36.03,535.45) -- cycle ;
\draw   (641.22,437.01) -- (936.16,437.01) -- (936.16,536.26) -- (641.22,536.26) -- cycle ;

\draw (6,77.78) node [anchor=north west][inner sep=0.75pt]  [rotate=-270] [align=left] {Input};
\draw (6,200) node [anchor=north west][inner sep=0.75pt]  [rotate=-270] [align=left] {Monodepth2};
\draw (6,290.98) node [anchor=north west][inner sep=0.75pt]  [rotate=-270] [align=left] {H-Net};
\draw (6,420.19) node [anchor=north west][inner sep=0.75pt]  [rotate=-270] [align=left] {ChiTransformer};
\draw (6,500.64) node [anchor=north west][inner sep=0.75pt]  [rotate=-270] [align=left] {Ours};

\end{tikzpicture}
}
\caption{\rev{\textbf{Qualitative results on the KITTI Eigen Split test set.} 
Our model produces finer details, particularly for objects at close to mid-range distances. 
This can be seen in the left column, where the traffic sign, windows, and doors are clearly preserved; 
in the middle column, where the house, tree, and post are better delineated; 
and in the right column, where the iron rail is reconstructed without incorrectly filling its gaps, unlike competing methods. 
Comparison images from other approaches were obtained from the respective authors' papers.
}}
\label{fig:66}
\end{figure*}

\subsection{DrivingStereo Dataset}
 
We also use the DrivingStereo dataset~\citep{yang2019drivingstereo}, which contains 174,437 stereo pairs for training collected over 38 video sequences and 7,751 stereo pairs collected in 4 video sequences for testing. DrivingStereo contains more diverse scenarios than KITTI, including urban, suburban, highway, elevated, and country roads, with sunny, rainy, cloudy, foggy, and dusky weather. \rev{In addition, it provides a 2,000-image subset specifically divided by weather conditions (sunny, cloudy, foggy, and rainy), which we use for further evaluation.}

\subsection{Estimating Depth Procedures}

We followed the procedures proposed in Monodepth2~\citep{godard2019digging} for the conversions of depth and disparity for the KITTI dataset training. We convert the network's sigmoid output $\sigma$ into depth $D$ by defining $D=1/(a \sigma + b)$. Since $\left \{ \sigma \in \mathbb{R}: 0< \sigma < 1  \right \}$, the $a$ and $b$ terms account for the conversion of the $\left (0, 1\right )$ range to a maximum and minimum disparity values based on a minimum and maximum depth of 0.1 and 100 units. Hence, since $B$ and $f_x$ are constants in the depth-disp conversion formula, the equivalence of both depth functions gives us that $\text{disp} = Bf_x(a \sigma + b)$. Therefore, the real disparity in pixels can be extracted by accounting for the predicted rescaled disparity extracted from $(a\sigma + b)$ multiplied by a $Bf_x$ term. 

During training, Monodepth2 also changed the translation in the $x$ direction of the extrinsic parameter to 0.1 instead of 0.54. Therefore, we must multiply the final depth predictions by $5.4$ to generate real depth results. The minimum and maximum depth predicted by our KITTI network is between 0.54 and 540 meters.

For the DrivingStereo dataset experiments, we also used the equation $\text{depth} = Bf_x /{\text{disp}}$, and since the baseline distance is the same as KITTI ($B=5.4$), we have a maximum and minimum depth of $0.54$ to $540$ meters. We also performed experiments with lower intervals in the KITTI dataset, but they did not improve the results.

\section{Results and Discussions}
 
\subsection{KITTI Results}
\label{sec:kitti}

\textbf{Comparison with Depth Estimation  Techniques:} metric results for the Eigen Split on both KITTI and IGT datasets are available in~\Cref{tab:1}, and a qualitative comparison is shown in Figure~\ref{fig:66}.
 For the IGT dataset, CCNeXt achieves state-of-the-art results in all metrics, surpassing all other reported monocular, monocular-stereo, and stereo self-supervised models. For the original Eigen Split, our approach achieves state-of-art results in the AbsRel, SqRel, and $\delta < 1.25$ metrics. ChiTransformer achieves state-of-the-art results for the other metrics, but it is important to highlight that this technique utilizes a ResNet-50 in the encoder combined with six Transformer-based self-attention mechanisms and six rectification blocks. Each rectification block contains two extra self-attention layers and one cross-attention layer. Thus, our approach is 10.18$\times$ faster than ChiTransformer, and achieves state-of-the-art metrics in the KITTI dataset, as displayed in~\Cref{tab:time}. To measure time, we ran each model 200 times using identical images from the testing set with batch size 1 and calculated the average and standard deviation for each execution. In instances where the code was unavailable, we replicated the architecture implementation based on information extracted from the papers to extract runtime. In~\Cref{tab:1}, ES$^3$Net's results were reported in this paper but ES$^3$Net was trained using the whole KITTI Raw, which comprises Eigen's training and testing data as part of the training dataset. This potentially favors their metrics by breaking the equality of comparison with other papers.

 \begin{table}[!htb]
\caption{\textbf{Runtime was measured for state-of-the-art self-supervised stereo techniques. 
The ChiTransformer architecture was reproduced based on the description provided in the original paper. 
\rev{Although we could not replicate the exact training procedure, the reproduced model based on their official repository is sufficient for a fair runtime estimation. 
All measurements were conducted with an input resolution of $1280{\times}384$ for CCNeXt and $1216{\times}352$ for ChiTransformer, 
using an Intel Xeon Gold 6148 CPU.
}
}}
\setlength{\tabcolsep}{13pt}
\centering
\label{tab:time}

\resizebox{\columnwidth}{!}{%
\begin{tabular}{ c  c  c }
\hline 

Method  &  \multicolumn{1}{c}{ChiTransformer~\citep{su2022chitransformer}} & \multicolumn{1}{c}{CCNeXt (ours)} \\

\hline 
Runtime(s) &  $29.82\pm1.00$ & $2.93\pm0.03$  \\

\hline 

\end{tabular}
}
\end{table}

\begin{table}[!t]
\caption{\textbf{Median Value Metric Comparison of Available Techniques on KITTI Eigen Split.} \dag $p<0.05$ for Wilcoxon test using Bonferroni correction.}
\setlength{\tabcolsep}{8pt}
\centering
\label{tab:stat}

\resizebox{\columnwidth}{!}{%
\begin{tabular}{ c  c  c  c  c }

\hline

 Method  & \multicolumn{1}{c}{AbsRel} & \multicolumn{1}{c}{SqRel} & \multicolumn{1}{c}{RMSE} & \multicolumn{1}{c}{$\delta < 1.25$} \\

\hline

H-Net~\citep{huang2022h}   & 0.072 & 0.489 & 3.830 & 0.927 \\

ES$^3$Net~\citep{fang2023es3net}   & 0.067 & 0.491 & 3.841  & 0.928 \\

\hline
CCNeXt (ours)   & \textbf{0.064\dag} & \textbf{0.445\dag} & \textbf{3.695\dag} & \textbf{0.933\dag}\\
\hline

\end{tabular}
}
\end{table}

Besides the highlighted differences in some metric results, \textit{e.g.}, in the AbsRel metric, it is possible to see many equal or similar results with differences lower than 5\%, which is not enough to imply a statistical gap between the outcomes of different methods. Furthermore, it is pertinent to note that while the average is frequently employed for method comparison, the distributions of the results presented here are non-parametric, as evidenced by a significance level of $p<0.05$ in the Shapiro-Wilk test. Consequently, utilizing the median in this context provides a more appropriate and robust representation.
As seen in~\Cref{tab:stat}, for $\delta < 1.25^2$ and $\delta < 1.25^3$, CCNeXt exhibits no discernible statistical difference when juxtaposed with other state-of-the-art methods. However, when analyzing AbsRel, SqRel, RMSE, and $\delta < 1.25$, CCNeXt demonstrates superior performance, evidenced by statistical significance ($p<0.05$ in the Wilcoxon test, with the application of the Bonferroni correction to account for multiple comparisons). Consequently, in comparison with H-Net and ES$^3$Net, our approach demonstrates superiority in four out of seven metrics while achieving similar performance in two metrics ($\delta < 1.25^2$ and $\delta < 1.25^3$) and exhibiting inferiority in just one (RMSE log).
While we intended to perform statistical analysis using our technique alongside the top three methods from the literature, there were no reproducible codes or weights available to replicate the results of some papers. As a result, only the H-Net and ES$^3$Net were employed in this phase.

\textbf{Comparison with Stereo Matching Techniques:} We performed experiments using supervised stereo-matching techniques since they had already been adapted to predicting disparity (consequently depth) in a stereo camera set. We conducted experiments using the IGEV~\citep{xu2023iterative} and PMN~\citep{chang2018pyramid} models applying ES$^3$Net~\citep{fang2023es3net} training procedures. The results are available in Table~\ref{tab:unsp}. \rev{To ensure a fair comparison, we report the performance of our model trained at a similarly low resolution, consistent with the settings used for PMN and IGEV.} PMN and IGEV did not achieve competitive results in this setting compared to our results. This is likely due to the fact that these architectures are typically optimized to enhance detailed predictions under a supervised loss function, often through complex decoders with various cost volume compositions.

\begin{table*}[!htb]
\caption{\textbf{Unsupervised Stereo Matching vs CCNeXt.} Results on the KITTI Eigen Split comparing the IGEV under self-supervised setting and CCNeXt. All stereo matching models were trained under self-supervised settings by us, since original papers only present results for supervised settings. Best results are \textbf{bolded}.}

\setlength{\tabcolsep}{6pt}
\centering
\label{tab:unsp}

\resizebox{\textwidth}{!}{%
\begin{tabular}{ c  c  c  c  c  c  c  c  c  c }
\toprule
\multirow{2}{*}{Method} & \multirow{2}{*}{Train} & \multirow{2}{*}{Input} & \multicolumn{4}{c}{\textit{lower is better}} & \multicolumn{3}{c}{\textit{higher is better}} \\ \cmidrule(lr){4-7} \cmidrule(lr){8-10}

 &  &  & \multicolumn{1}{c}{AbsRel} & \multicolumn{1}{c}{SqRel} & \multicolumn{1}{c}{RMSE} &  \multicolumn{1}{c}{RMSE log} & \multicolumn{1}{c}{$\delta < 1.25$} & \multicolumn{1}{c}{$\delta < 1.25^2$} & \multicolumn{1}{c}{$\delta < 1.25^3$}\\

\hline 

PMN~\citep{chang2018pyramid} (Self-Supervised Training) & S & LR & 1.659  &  45.910  &  20.786  &   0.986  &   0.155  &   0.301  &   0.440  \\

IGEV~\citep{xu2023iterative} (Self-Supervised Training) &  S  & LR &  0.187  &   5.372  &   7.251  &   0.331  &   0.889  &   0.926  &   0.946  \\


CCNeXt (ours)  & S  & LR & \textbf{0.075} & \textbf{0.639} & \textbf{4.140} & \textbf{0.175} & \textbf{0.920} & \textbf{0.962} & \textbf{0.979} \\
\bottomrule

\end{tabular}
}
\end{table*}


 \subsection{DrivingStereo Results}

Due to the reduced number of evaluation images and low variable scenarios of the KITTI dataset, we perform experiments on a more complex dataset. For that reason, we chose the DrivingStereo dataset. We intended to compare all self-supervised depth estimation published techniques for this evaluation, but unfortunately, we only have training code that allows us to adapt the techniques for a new dataset for ES$^3$Net and older techniques such as Monodepth2.











\begin{table}[!htb]
\caption{\textbf{Depth Metric Results on the DrivingStereo.} Comparative results with ES$^3$Net using depth estimation metrics. The best results are in \textbf{bold}.}
\setlength{\tabcolsep}{4pt}
\centering
\label{tab:2}

\resizebox{\columnwidth}{!}{%
\begin{tabular}{ c c c c c }

\toprule

\multirow{2}{*}{Method} & \multicolumn{3}{c}{\textit{lower is better}} & \multicolumn{1}{c}{\textit{higher is better}} \\ 

\cmidrule(lr){2-4} \cmidrule(lr){5-5}

   & \multicolumn{1}{c}{AbsRel} & \multicolumn{1}{c}{SqRel} & \multicolumn{1}{c}{RMSE} & \multicolumn{1}{c}{$\delta < 1.25$} \\

\midrule
Monodepth2~\citep{godard2019digging} & 0.169  &   3.669  &   10.283  &  0.805  \\

ES$^3$Net~\citep{fang2023es3net} & 0.208  &   10.917  &   11.206  &  0.885  \\

CCNeXt (ours)  & \textbf{0.067} & \textbf{1.270} & \textbf{5.457} & \textbf{0.9619} \\
\bottomrule
\end{tabular}
}
\end{table}

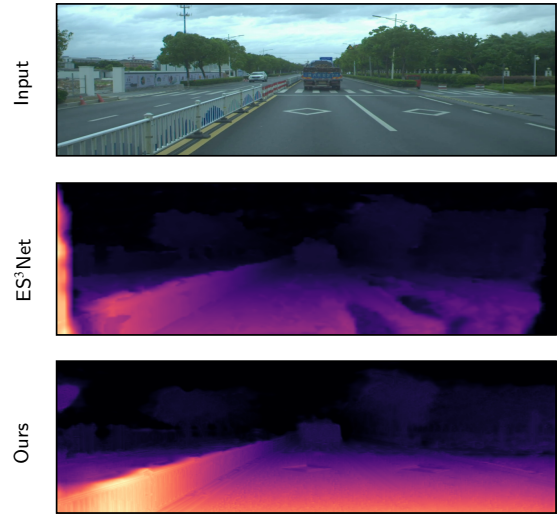
\begin{figure}[!htb]
\begin{minipage}{\columnwidth}
\resizebox{0.9\textwidth}{!}{
\centering

\tikzset{every picture/.style={line width=0.75pt}} 

\begin{tikzpicture}[x=0.75pt,y=0.75pt,yscale=-1,xscale=1]

\draw (261.08,275.53) node  {\includegraphics[width=222.99pt,height=67.81pt]{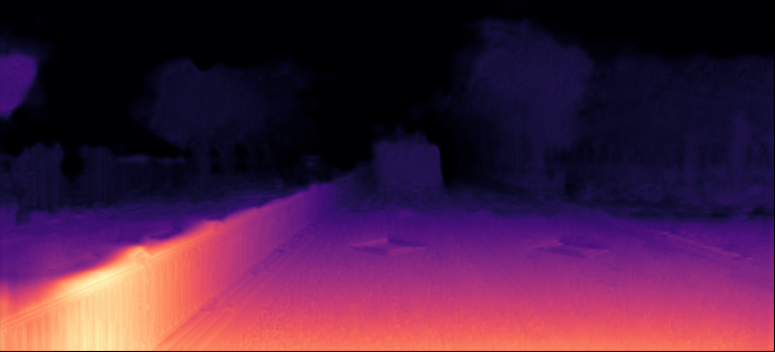}};
\draw (261.08,62.86) node  {\includegraphics[width=222.99pt,height=67.81pt]{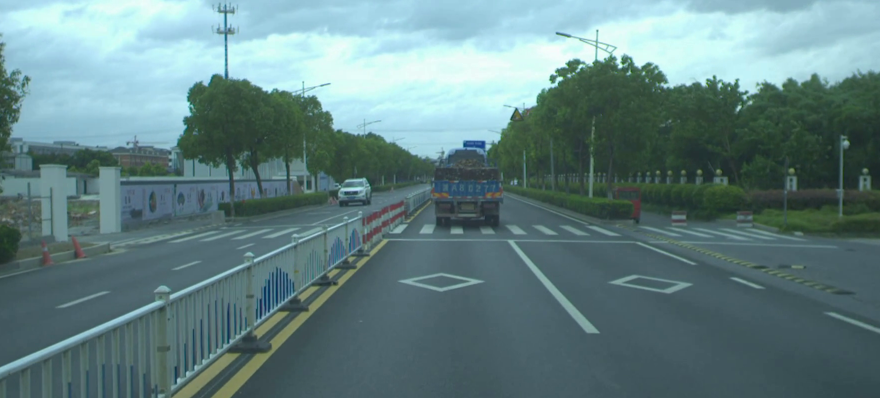}};
\draw (261.08,169.53) node  {\includegraphics[width=222.99pt,height=67.81pt]{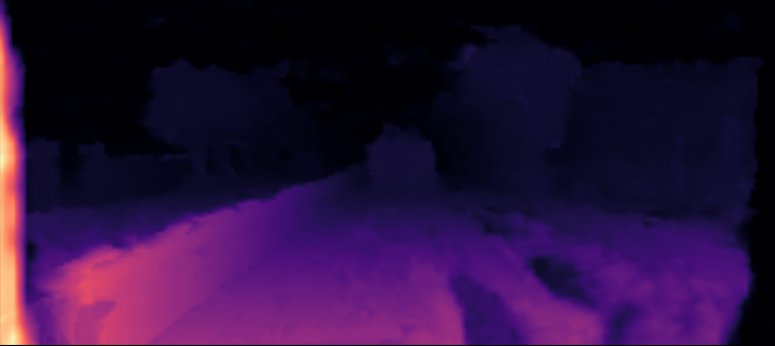}};
\draw   (112.42,17.65) -- (409.74,17.65) -- (409.74,108.07) -- (112.42,108.07) -- cycle ;
\draw   (112.42,230.32) -- (409.74,230.32) -- (409.74,320.74) -- (112.42,320.74) -- cycle ;
\draw   (112.42,124.32) -- (409.74,124.32) -- (409.74,214.74) -- (112.42,214.74) -- cycle ;

\draw (85.68,79.85) node [anchor=north west][inner sep=0.75pt]  [rotate=-270] [align=left] {Input};
\draw (85.68,293.03) node [anchor=north west][inner sep=0.75pt]  [rotate=-270] [align=left] {Ours};
\draw (85.68,196.53) node [anchor=north west][inner sep=0.75pt]  [rotate=-270] [align=left] {ES$^3$Net};

\end{tikzpicture}

}
\end{minipage}
  \caption{On \textbf{top}: Left-view input image from the DrivingStereo dataset. On \textbf{middle}: Disparity estimation of ES$^3$Net. On \textbf{bottom:} Results of our model (CCNeXt).}
  
  \label{fig:9}

\end{figure}

\begin{table}[!htb]
\setlength{\tabcolsep}{2mm}
\centering
\small
\caption{\textbf{DrivingStereo Results.} EPE and D1 error metrics reported on the DrivingStereo dataset. In the \textit{Train} column: \textbf{D}: supervised stereo-matching and \textbf{S}: stereo pair depth self-supervised.}
\label{tab:2ster}

\resizebox{\columnwidth}{!}{%
\begin{tabular}{c c c c }

\toprule 

 Method  & Train & \multicolumn{1}{c}{ EPE } & \multicolumn{1}{c}{D1 error} \\
\midrule 

SegStereo~\citep{yang2018segstereo} & D & 1.13 & 5.31\% \\

EdgeStereo~\citep{song2020edgestereo} & D & 1.12 & 5.35\% \\

GuideNet~\citep{yang2019drivingstereo} & D & 1.36 & 7.33\% \\

\hline 
  ES$^3$Net~\citep{fang2023es3net} & S    & 4.406 & 16.86\% \\

 CCNeXt (ours)  & S & \textbf{1.45} & \textbf{6.5\%} \\
\bottomrule

\end{tabular}
}
\end{table}

We report depth metrics in~\Cref{tab:2} and present qualitative results in~\Cref{fig:9}. Comparing the performance of CCNeXt with ES$^3$Net and Monodepth2 on the DrivingStereo dataset, there is a reduction of $67.79\%$ and $60.36\%$ for AbsRel, $88.37\%$ and $65.38\%$ for SqRel, and $51.30\%$ and $46.93\%$ for RMSE, respectively. These reductions are significantly larger than the respective reductions of $5.4\%$ and $34.58\%$ for AbsRel, $11.97\%$ and $31.57\%$ for SqRel, $3.93\%$ and $18.49\%$ for RMSE on the KITTI dataset. Thus, it is noteworthy to assert that evaluating depth on the DrivingStereo dataset appears to accentuate the disparities between the two models. This occurs due to the increased complexity of the DrivingStereo dataset compared to KITTI.

We also report EPE and D1 error metrics in Table~\ref{tab:2ster}, since the DrivingStereo dataset presents disparity ground truth results in the original paper~\citep{yang2019drivingstereo}. Comparing the performance of CCNeXt with ES$^3$Net, there is a 67.09\% and 61.45\% reduction in the EPE and D1 error metrics, respectively. We also included a comparison with other supervised reported models, which shows that the CCNeXt also achieves close metric errors compared to supervised models, even though our method solely relies on the stereo pair during training. EdgeStereo~\citep{song2020edgestereo} presents 22.75\% and 17.69\% reduction compared to CCNeXt, but it is a supervised strategy.

\rev{Additionally, We evaluate our method on the weather subset, comparing against ES$^3$Net and MonoDepth2 and results are available in Table~\ref{tab:weather}. Our model performs consistently across Cloudy, Foggy, and Sunny conditions, despite not being explicitly trained for weather robustness. It achieves the best results in 27 of 28 metrics, with performance dropping mainly in Rainy, followed by Foggy condition. ES$^3$Net shows a similar trend, with larger differences between Sunny and Cloudy. Although MonoDepth2 outperforms ES$^3$Net on the default DrivingStereo test set, it performs worse on this weather split, except in Rainy, where it achieves its best results. Its ranking across conditions (Rainy > Foggy > Sunny > Cloudy) contrasts with the more stable behavior of CCNeXt and ES$^3$Net.}

\rev{For weather splits, CCNeXt reduces Abs Rel, Sq Rel, and RMSE by 69.86\%, 77.14\%, and 46.97\% when compared to the second best solution, close to the reductions on the full test set (60.36\%, 65.38\%, and 46.93\%, respectively). The Rainy results highlight the need for strategies that explicitly address weather generalization. Our training relied only on basic augmentations (brightness, hue, contrast), similar to MonoDepth2, and incorporating rain-specific augmentations (e.g., raindrop effects) could already improve performance.}

\begin{table*}[!htb]
\caption[Depth Metric Results on the DrivingStereo Weather Split.]{\rev{\textbf{Depth Metric Results on the DrivingStereo Weather Split.} Comparative results for the four different weather splits using Monodepth2 and ES$^3$Net. The best results are in \textbf{bold}.}}  
\setlength{\tabcolsep}{10pt}
\centering
\label{tab:weather}
\resizebox{\textwidth}{!}{%
\begin{tabular}{c c  c  c  c  c  c  c  c }
\toprule
Domain & \multirow{2}{*}{Method} & \multicolumn{4}{c}{\textit{lower is better}} & \multicolumn{3}{c}{\textit{higher is better}} \\ \cmidrule(lr){3-6} \cmidrule(lr){7-9}

& &  \multicolumn{1}{c}{Abs Rel} & \multicolumn{1}{c}{Sq Rel} & \multicolumn{1}{c}{RMSE} &  \multicolumn{1}{c}{RMSE log} & \multicolumn{1}{c}{$\delta < 1.25$} & \multicolumn{1}{c}{$\delta < 1.25^2$} & \multicolumn{1}{c}{$\delta < 1.25^3$}\\

\midrule 
 
\multirow{3}{*}{\rotatebox{90}{Cloudy}} &  

MonoDepth2~\citep{godard2019digging} & 0.412 & 14.953 & 15.892 & 0.411 & 0.540 & 0.867 & 0.936 \\

& ES$^3$Net~\citep{fang2023es3net} & 0.187 & 10.766 & 9.429 & 0.289 & 0.906 & 0.940 & 0.957 \\

& CCNeXt (ours) & \textbf{0.052} & \textbf{0.668} & \textbf{4.388} & \textbf{0.105} & \textbf{0.971} & \textbf{0.989} & \textbf{0.995} \\

\hline 

\multirow{3}{*}{\rotatebox{90}{Foggy}} &

MonoDepth2~\citep{godard2019digging} & 0.345 & 11.141 & 16.850 & 0.361 & 0.567 & 0.892 & 0.950 \\

& ES$^3$Net~\citep{fang2023es3net} & 0.254 & 13.811 & 13.203 & 0.355 & 0.854 & 0.913 & 0.938 \\

& CCNeXt (ours) & \textbf{0.061} & \textbf{0.587} & \textbf{5.409} & \textbf{0.115} & \textbf{0.956} & \textbf{0.987} & \textbf{0.994} \\

\hline

\multirow{3}{*}{\rotatebox{90}{Sunny}} &

MonoDepth2~\citep{godard2019digging} & 0.407 & 14.395 & 15.509 & 0.419 & 0.555 & 0.855 & 0.928 \\

& ES$^3$Net~\citep{fang2023es3net}  & 0.148 & 6.590 & 8.104 & 0.273 & 0.902 & 0.942 & 0.960 \\

& CCNeXt (ours)  & \textbf{0.054} & \textbf{0.569} & \textbf{4.279} & \textbf{0.112} & \textbf{0.963} & \textbf{
0.987} & \textbf{0.994} \\

\hline

\multirow{3}{*}{\rotatebox{90}{Rainy}} &  

MonoDepth2~\citep{godard2019digging} & 0.281 & 5.306 & 13.860 & 0.306 & 0.599 & 0.896 & 0.959 \\

& ES$^3$Net~\citep{fang2023es3net} & 0.570 & 36.889 & 25.986 & 0.756 & 0.668 & 0.745 & 0.788 \\

& CCNeXt (ours)  &  \textbf{0.118} & \textbf{3.398} & \textbf{7.623} & \textbf{0.197} & \textbf{0.927} & \textbf{0.966} & \textbf{0.978} \\

\bottomrule

\end{tabular}
}
\end{table*}

\subsection{Computational Efficiency (FLOP Analysis)}

\rev{In addition to runtime comparisons between our model and ChiTransformer, we conduct a detailed analysis of computational complexity in terms of floating\mbox{-}point operations (FLOPs). 
FLOPs are computed using \texttt{fvcore} \texttt{FlopCountAnalysis} with an input resolution of $1280{\times}384$ for CCNeXt and $1216{\times}352$ for ChiTransformer, both with a batch size of $1$, under a consistent counting convention across models. 
The analysis includes convolutional, linear, and attention layers, while excluding element-wise operations such as activations and normalizations, following standard practice. 
Table~\ref{tab:flops} summarizes the total FLOPs for both the full architectures and their respective encoder components.
}
\begin{table}[h]
\centering
\footnotesize
\caption{\rev{\textbf{Comparison of FLOPs (GigaFLOPs) Measured Using \texttt{fvcore} \texttt{FlopCountAnalysis}.} Full architectures and encoders are computed at $1280{\times}384$ for CCNeXt and at $1216{\times}352$ for ChiTransformer, batch size $1$. 
Per-layer cross-attention FLOPs are measured on $(1,64,144,320)$, matching our first CA stage.}}

\begin{tabular*}{\columnwidth}{@{\extracolsep{\fill}} l c}

\toprule
Model / component & FLOPs (G) \\
\midrule
\multicolumn{2}{l}{\textit{Full Architecture}} \\
ChiTransformer~\citep{su2022chitransformer} & 665 \\
CCNeXt (ours) & 65.942 \\
\midrule
\multicolumn{2}{l}{\textit{Encoder Only}} \\
ChiTransformer~\citep{su2022chitransformer} & 68.168 \\
CCNeXt (ours) & 40.264 \\
\midrule
\multicolumn{2}{l}{\textit{Per-layer Cross-Attention (on $(1,64,144,320)$)}} \\
Full-row CA & 3.563 \\
Windowed Epipolar CA (ours) & 1.046 \\
\bottomrule
\end{tabular*}
\label{tab:flops}
\end{table}

\rev{When isolating the encoder stage, ChiTransformer exhibits a computational cost of 68.168 GFLOPs, while our CCNeXt encoder requires only 40.264 GFLOPs. This difference becomes even more pronounced when comparing the entire architectures: ChiTransformer demands 665 GFLOPs, whereas CCNeXt operates at just 65.942 GFLOPs, an order of magnitude lower. These results highlight that the extensive use of self-attention and cross-attention layers in ChiTransformer significantly increases its computational burden. In contrast, our architecture maintains a lightweight decoder, which accounts for only 39\% of the total FLOPs.}

\rev{A key contributor to this efficiency is our Windowed Cross-Attention mechanism. 
To quantify its impact, we compare it against standard full cross-attention. 
For an input of shape $(1, 64, 144, 320)$, corresponding to the dimensions entering the first cross-attention layer in our network, the full cross-attention module incurs 3.563 GFLOPs, whereas our windowed variant requires only 1.046 GFLOPs, representing a 70.6\% reduction in computational cost. 
Given that multiple cross-attention layers are used throughout our encoder, this saving accumulates across the network, making CCNeXt a favorable option when balancing accuracy with computational efficiency.}

\section{Ablation Studies}

This section presents ablation studies to analyze the contribution of individual stages to our final results (\Cref{tab:33}). \rev{We first evaluate changes incrementally, and then compare alternative encoder backbones.}

\begin{table}[!htb]
\setlength{\tabcolsep}{2pt}
\centering
\caption{\textbf{Ablation Studies.} 
Evaluation of the ConvNeXt encoder, the impact of windowed cross-attention (W-CA), and the contributions of the ICEP module with skip blocks on the KITTI Eigen Split dataset. 
Rows labeled “Monodepth2 decoder” correspond to the decoder from~\citet{godard2019digging}. 
For comparison, we also include two Transformer-based encoders alongside ConvNeXt.}

\resizebox{\columnwidth}{!}{%
\begin{tabular}{lcccc}
\toprule
\multirow{2}{*}{Method} & \multicolumn{3}{c}{\textit{lower is better}} & \textit{higher is better} \\ 
\cmidrule(lr){2-4} \cmidrule(lr){5-5}
& AbsRel & SqRel & RMSE & $\delta < 1.25$ \\
\midrule

\multicolumn{5}{l}{\textbf{Incremental Ablation Steps}} \\
ResNet encoder + Monodepth2 decoder &   0.086  &   0.631  &   4.192  & 0.914   \\
ConvNeXt encoder (no W\mbox{-}CA) + Monodepth2 decoder &   0.077  &   0.639  &   4.187  &   0.922   \\
CCNeXt encoder (with W\mbox{-}CA) + Monodepth2 decoder & 0.072  &   0.654  &   4.176  & 0.925 \\
CCNeXt encoder (with W\mbox{-}CA) + CCNeXt decoder & \textbf{0.070}  & \textbf{0.581} & \textbf{3.883} & \textbf{0.926} \\
\midrule

\multicolumn{5}{l}{\textbf{Additional Encoder Experiments}} \\
Swin Transformer Encoder      & 0.075 & 0.617 & 4.077 & 0.924 \\
PVT Encoder                   & 0.072 & 0.583 & 3.995 & 0.923 \\
ConvNeXt Encoder & \textbf{0.070}  & \textbf{0.581} & \textbf{3.883} & \textbf{0.926} \\

\bottomrule
\end{tabular}
}
\label{tab:33}
\end{table}



\textbf{Encoder.}  
When comparing the ResNet-based encoder with the ConvNeXt without the cross-attention layer, it becomes clear that ConvNeXt consistently provides more reliable feature representations for depth prediction. The improvement is most noticeable in the AbsRel metric, where ConvNeXt reduces the relative error, but its benefits also extend to more stable predictions across other metrics. This suggests that the architectural refinements in ConvNeXt, such as larger receptive fields and improved convolutional blocks, contribute to better capturing mid- and long-range dependencies that are important for dense regression tasks like depth estimation.  
Moreover, once the windowed cross-attention is incorporated, the results further improve, particularly by lowering the RMSE and boosting $\delta < 1.25$. This indicates that the attention mechanism enables the encoder to refine contextual relationships across spatial regions, which are crucial for scenes containing multiple objects at varying scales and distances.

\textbf{Decoder.}  
The decoder design also plays a critical role in the final performance. Compared to the baseline Monodepth2 decoder, our proposed decoder combined with skip connections produces consistent improvements across nearly all metrics. The most significant impact appears in the SqRel and RMSE metrics, where the inclusion of Skip blocks reduces large squared errors. This effect can be attributed to the Skip connections mitigating phantom regions and enforcing better spatial consistency (\Cref{fig:5}). By preserving low-level details from earlier layers and mixing them with high-level semantic features, the decoder ensures smoother depth transitions, particularly at object boundaries. Additionally, the ICEP module enhances the information exchange between encoder and decoder stages, enabling the network to better reconstruct fine-grained details such as thin structures (e.g., poles, railings, and traffic signs) while avoiding the over-smoothing often observed in standard decoders.

\begin{figure*}[!htb]
  \centering

\resizebox{0.9\textwidth}{!}{
\begin{tikzpicture}[x=0.75pt,y=0.75pt,yscale=-1,xscale=1]

\draw (486.32,486.64) node  {\includegraphics[width=221.2pt,height=74.44pt]{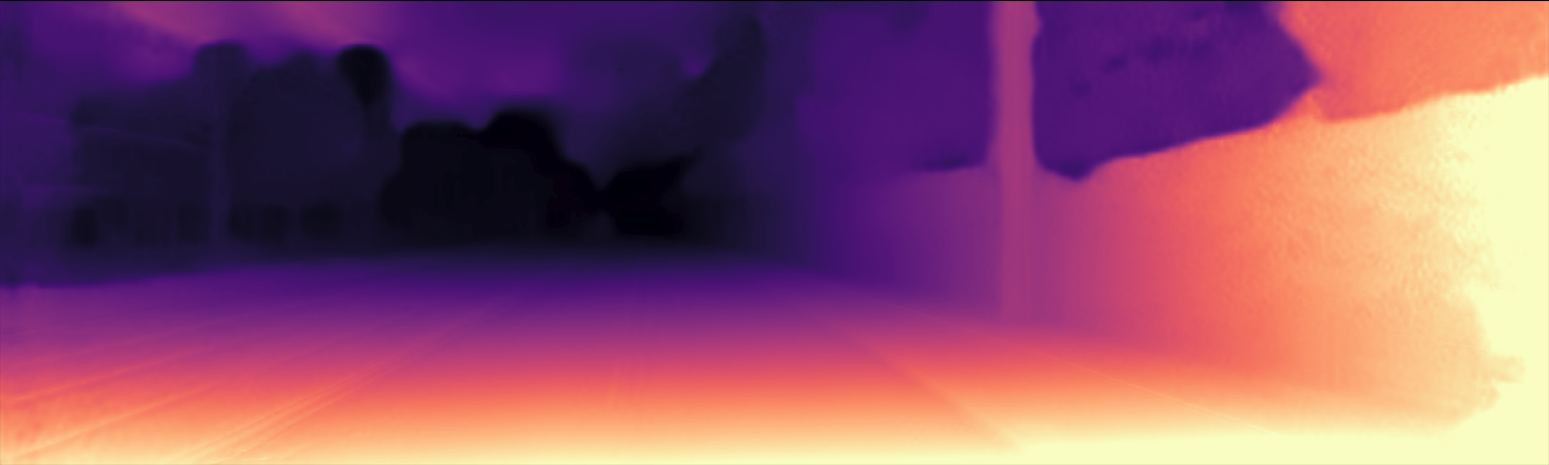}};
\draw (788.69,486.64) node  {\includegraphics[width=221.2pt,height=74.44pt]{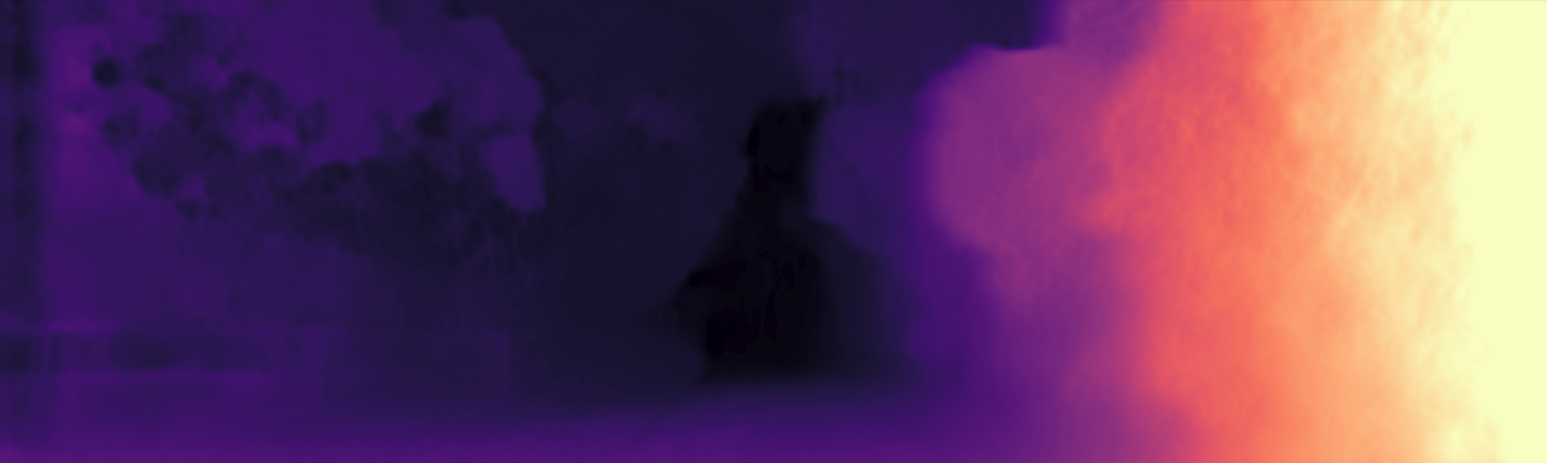}};
\draw (183.5,485.82) node  {\includegraphics[width=221.2pt,height=74.44pt]{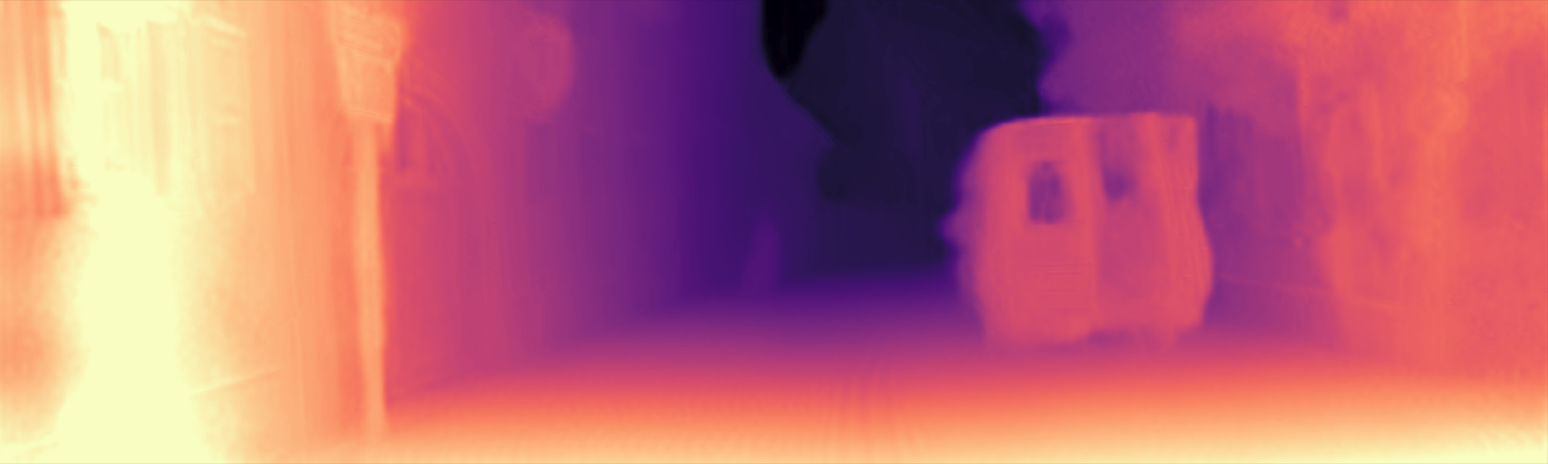}};
\draw (183.15,378.69) node  {\includegraphics[width=221.2pt,height=74.44pt]{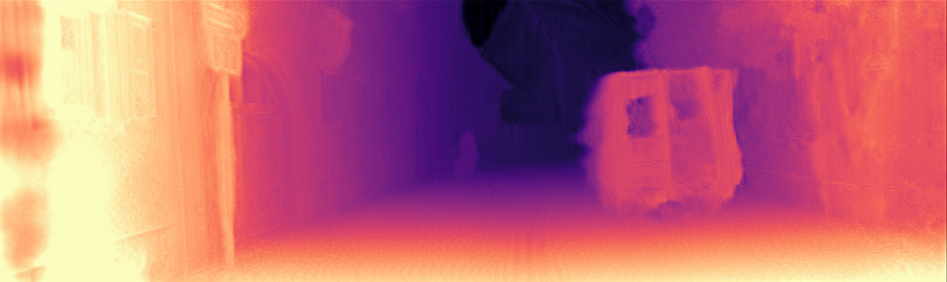}};
\draw (788.35,379.5) node  {\includegraphics[width=221.2pt,height=74.44pt]{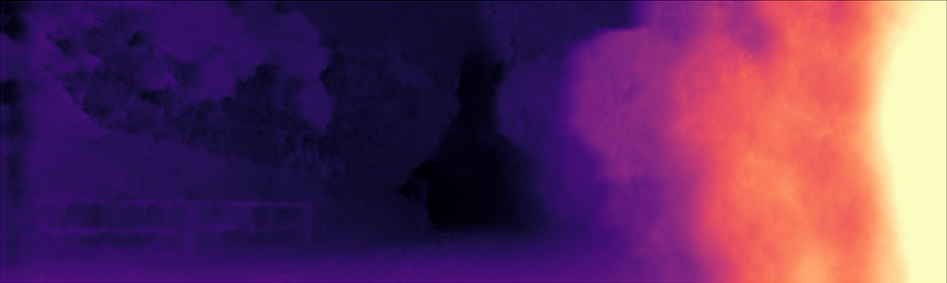}};
\draw (485.97,379.5) node  {\includegraphics[width=221.2pt,height=74.44pt]{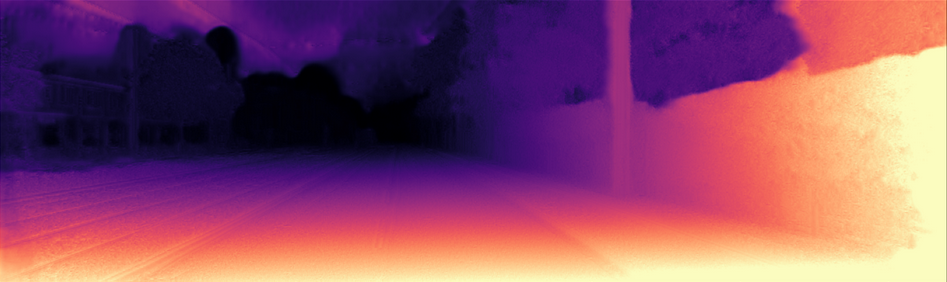}};
\draw (788.35,272.48) node  {\includegraphics[width=221.2pt,height=74.44pt]{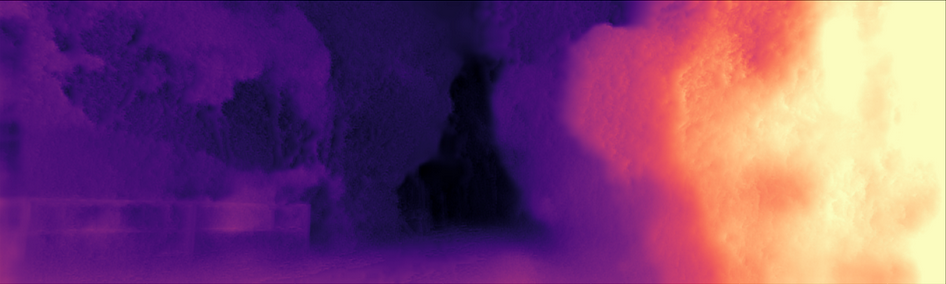}};
\draw (183.15,272.2) node  {\includegraphics[width=221.2pt,height=74.44pt]{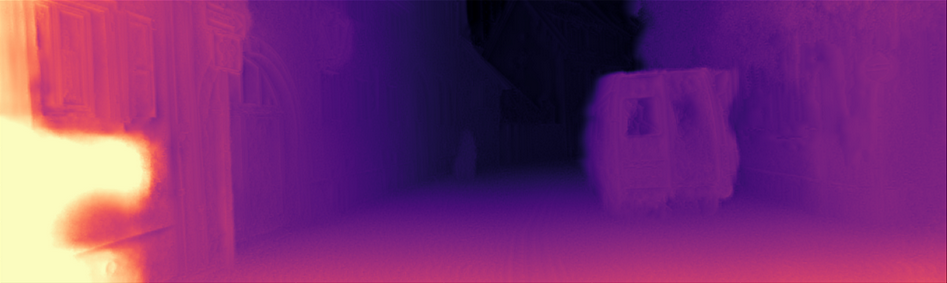}};
\draw (485.97,272.48) node  {\includegraphics[width=221.2pt,height=74.44pt]{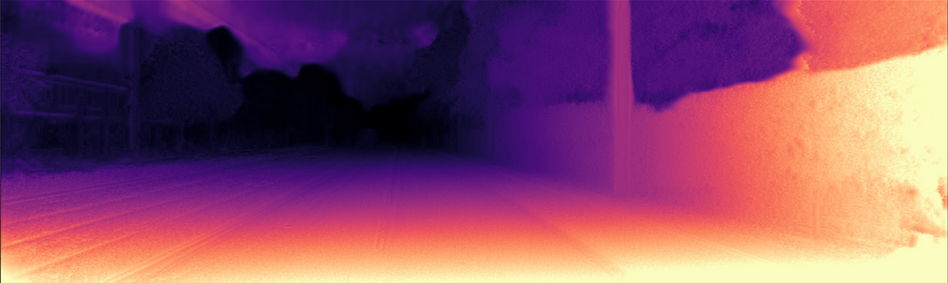}};
\draw (183.15,166.32) node  {\includegraphics[width=221.2pt,height=74.44pt]{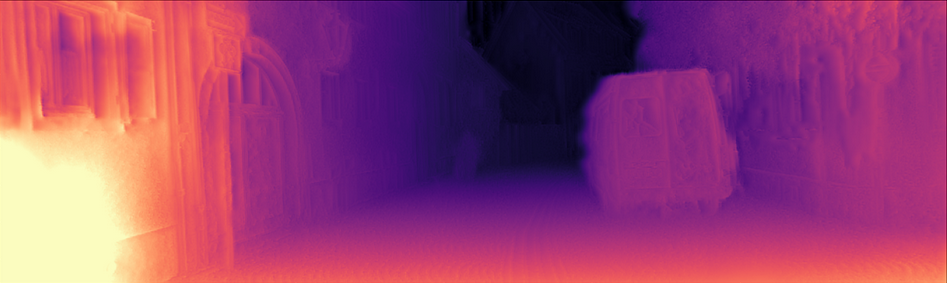}};
\draw (485.28,166.32) node  {\includegraphics[width=221.2pt,height=74.44pt]{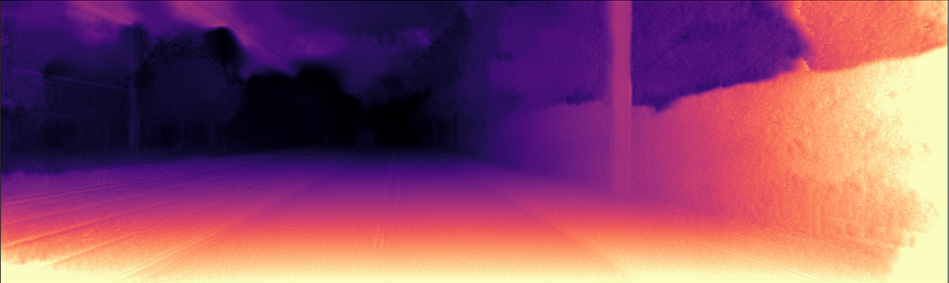}};
\draw (788.73,166.32) node  {\includegraphics[width=221.2pt,height=74.44pt]{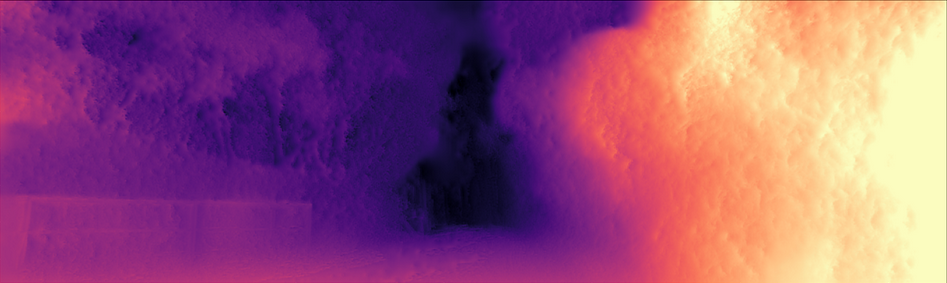}};
\draw (788.3,60.28) node  {\includegraphics[width=221.2pt,height=74.44pt]{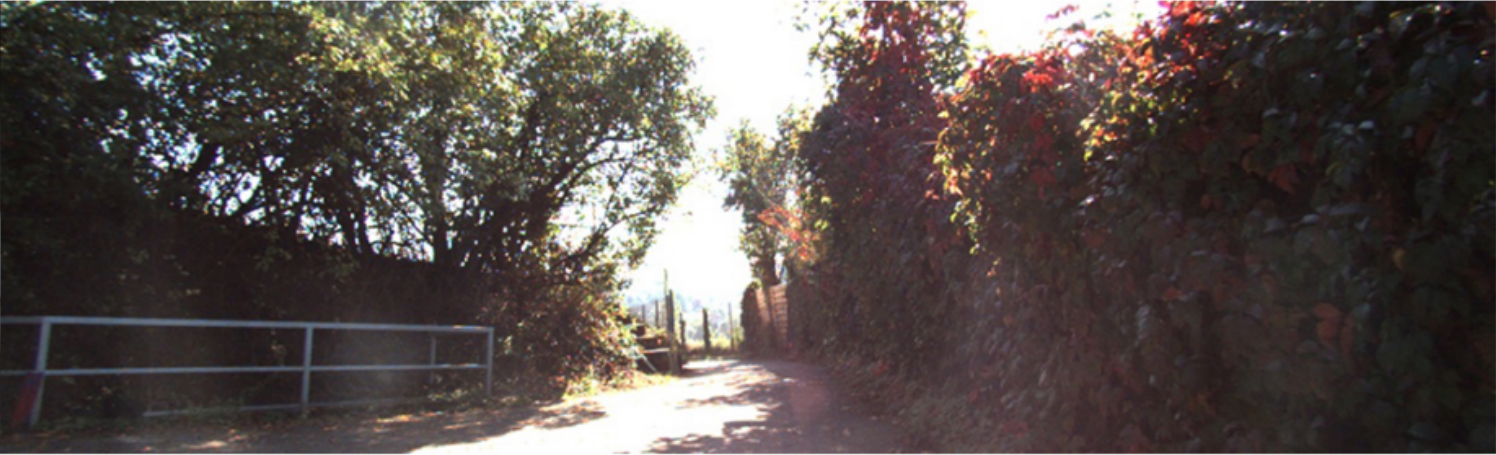}};
\draw (183.15,60.28) node  {\includegraphics[width=221.2pt,height=74.44pt]{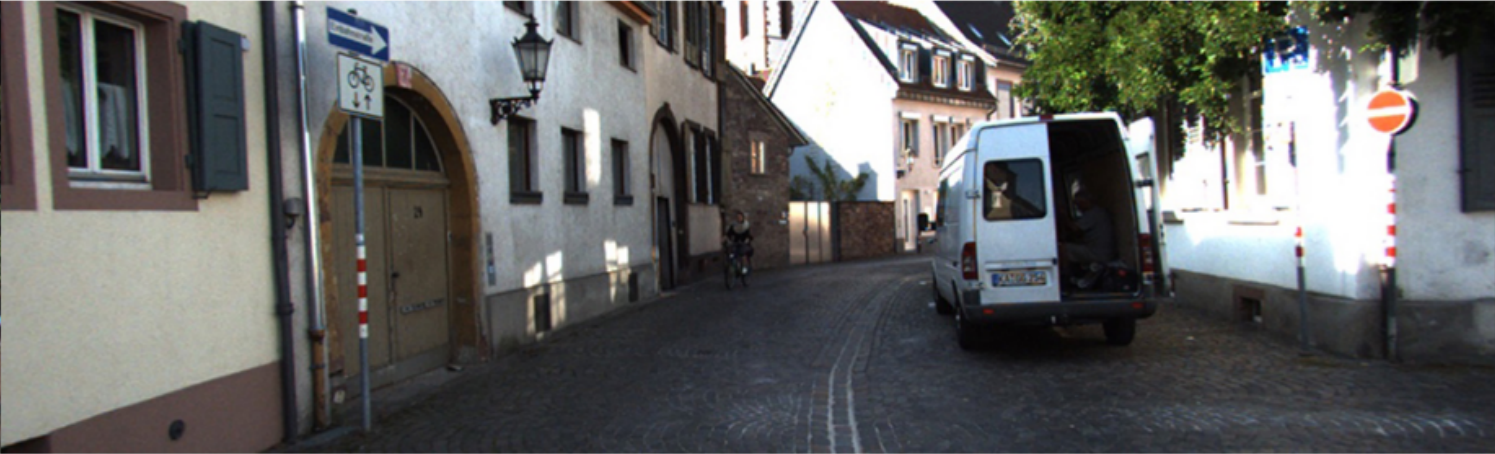}};
\draw (484.85,60.28) node  {\includegraphics[width=221.2pt,height=74.44pt]{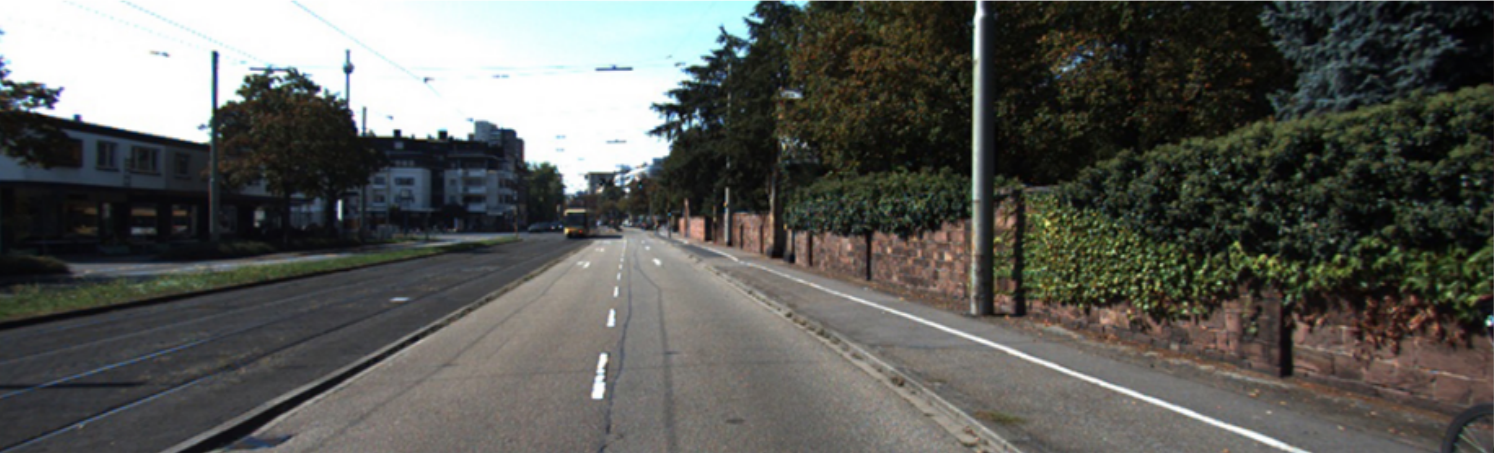}};
\draw   (337.82,116.69) -- (632.75,116.69) -- (632.75,215.95) -- (337.82,215.95) -- cycle ;
\draw   (338.5,222.85) -- (633.44,222.85) -- (633.44,322.1) -- (338.5,322.1) -- cycle ;
\draw   (338.5,329.87) -- (633.44,329.87) -- (633.44,429.13) -- (338.5,429.13) -- cycle ;
\draw   (35.68,329.06) -- (330.62,329.06) -- (330.62,428.31) -- (35.68,428.31) -- cycle ;
\draw   (35.68,222.58) -- (330.62,222.58) -- (330.62,321.83) -- (35.68,321.83) -- cycle ;
\draw   (35.68,116.69) -- (330.62,116.69) -- (330.62,215.95) -- (35.68,215.95) -- cycle ;
\draw   (35.68,10.65) -- (330.62,10.65) -- (330.62,109.91) -- (35.68,109.91) -- cycle ;
\draw   (337.38,10.65) -- (632.32,10.65) -- (632.32,109.91) -- (337.38,109.91) -- cycle ;
\draw   (641.26,116.69) -- (936.2,116.69) -- (936.2,215.95) -- (641.26,215.95) -- cycle ;
\draw   (640.88,222.85) -- (935.81,222.85) -- (935.81,322.1) -- (640.88,322.1) -- cycle ;
\draw   (640.88,329.87) -- (935.81,329.87) -- (935.81,429.13) -- (640.88,429.13) -- cycle ;
\draw   (640.83,10.65) -- (935.77,10.65) -- (935.77,109.91) -- (640.83,109.91) -- cycle ;
\draw   (338.85,437.01) -- (633.78,437.01) -- (633.78,536.26) -- (338.85,536.26) -- cycle ;
\draw   (36.03,436.19) -- (330.96,436.19) -- (330.96,535.45) -- (36.03,535.45) -- cycle ;
\draw   (641.22,437.01) -- (936.16,437.01) -- (936.16,536.26) -- (641.22,536.26) -- cycle ;

\draw (6,77.78) node [anchor=north west][inner sep=0.75pt]  [rotate=-270] [align=left] {Input};
\draw (6,199.32) node [anchor=north west][inner sep=0.75pt]  [rotate=-270] [align=left] {ResNet Enc.};
\draw (6,310.98) node [anchor=north west][inner sep=0.75pt]  [rotate=-270] [align=left] {ConvNeXt Enc.};
\draw (6,412.19) node [anchor=north west][inner sep=0.75pt]  [rotate=-270] [align=left] {CCNeXt Enc.};
\draw (6,510.64) node [anchor=north west][inner sep=0.75pt]  [rotate=-270] [align=left] {CCNeXt};

\end{tikzpicture}
}

\caption{\textbf{Qualitative results of the ablation models on the KITTI Eigen Split test set}. CCNeXt exhibits better consistency over region predictions than the other methods. It also produces less noisy regions.}
\label{fig:sup1}
\end{figure*}
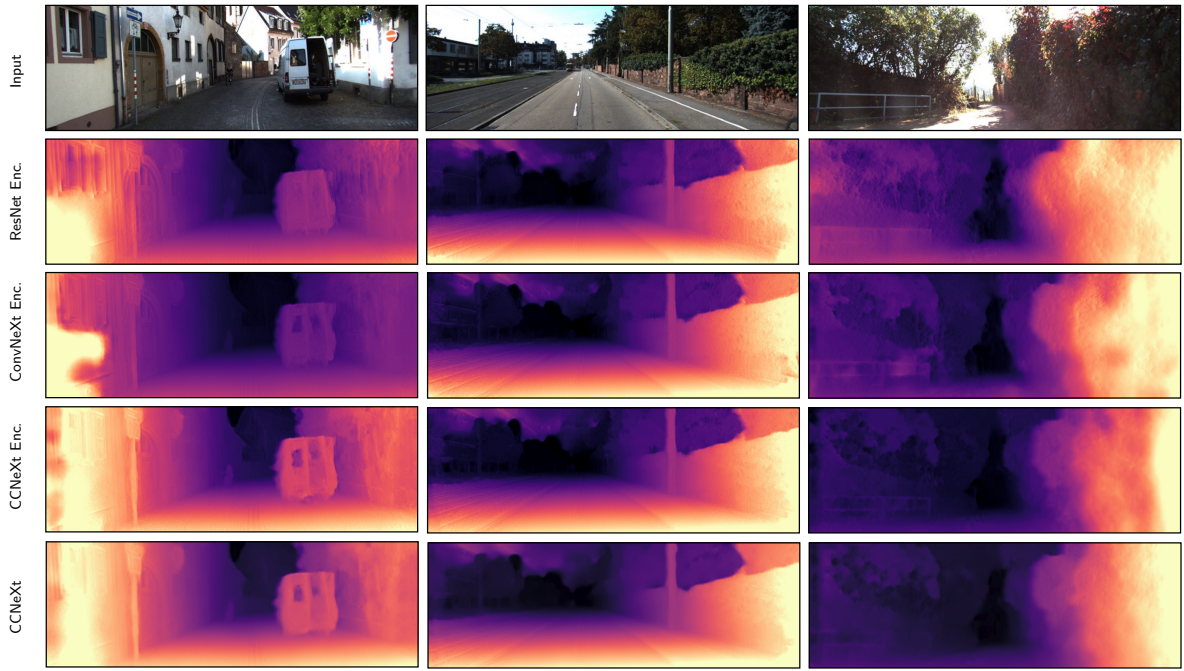

\rev{Qualitative results for these variants are shown in Figure~\ref{fig:sup1}. Relative to the ResNet baseline with the Monodepth2 decoder, the CCNeXt architecture produces smoother and more coherent depth maps, with fewer regions of large error. It also captures relative depth ordering more accurately (e.g., car vs. traffic sign, traffic sign vs. wall), leading to more consistent scene geometry.}

\rev{
\textbf{Additional Encoder Experiments.}  
To further validate the choice of ConvNeXt as our default encoder, we extended our analysis to two representative Transformer-based alternatives: the Swin Transformer~\citep{liu2021swin} and Pyramid Vision Transformer v2 (PVTv2)~\citep{wang2021pvtv2}. Both models were selected in configurations with computational costs close to our ConvNeXt variant, allowing for a fair comparison in terms of accuracy–efficiency trade-off. The tested Swin Transformer (Tiny version, complete encoder has 105 GFLOPs) and PVTv2 (PVTv2-B1, complete encoder has 47.052 GFLOPs) were compared against our ConvNeXt (complete encoder has 43.567 GFLOPs), where the Transformer-based baselines were further enhanced with 
Windowed Cross-Attention blocks between encoder stages to ensure a 
consistent comparison.}

\rev{
Performance remains below ConvNeXt across all major metrics, particularly in AbsRel and RMSE. This suggests that, in the context of the stereo self-supervised depth estimation, local window-based self-attention is less effective than ConvNeXt’s convolutional designs for capturing dense spatial correlations. Furthermore, the significantly higher FLOPs of Swin highlight an unfavorable accuracy–efficiency balance, making it less practical for real-world deployment.}

\rev{In contrast, the PVT encoder yields results that are close to ConvNeXt, with minor degradation in precision-based metrics such as $\delta < 1.25$. The hierarchical design of PVT appears to retain competitive representational power while remaining computationally reasonable. However, ConvNeXt still offers the best overall trade-off: it achieves the lowest AbsRel and SqRel, while maintaining lower complexity than both Transformer alternatives.}  

\rev{These findings underscore that, despite the recent popularity of Transformer-based encoders, well-designed convolutional architectures such as ConvNeXt remain highly competitive for dense regression problems, particularly when computational efficiency is a priority. In particular, ConvNeXt balances accuracy, efficiency, and stability across diverse depth estimation metrics, making it a strong choice for scenarios where computational budgets are constrained. Nonetheless, PVT emerges as a promising candidate for scaling to larger and more diverse datasets, where the global reasoning capacity of Transformers may provide further benefits.}

\section{Limitations and Future Directions}

Our model has three main limitations: performing inference with (i) different datasets, (ii) multi-dataset training, and (iii) non-rectified systems. Since we are performing the backward projection from 2D to 3D and then the perspective projection from 3D to 2D, we depend on the camera system's internal parameters to infer depth and create the reprojection image. Thus, performing inference in unseen datasets could be problematic, especially using the predicted rescaled disparity extracted from $(a\sigma + b)$.

This is also the same limitation encountered in training using multiple datasets simultaneously. We could solve the rescaling disparity limitation if we directly used equation $\text{depth} = Bf_x /{\text{disp}}$ as performed in the DrivingStereo setting. Another possible solution for both limitations would be to predict the pixel disparity with the model and directly warp one image view to generate the other using disparity information. In this case, it would be necessary to define minimum and maximum pixel disparity values as the same for all datasets.

\begin{figure}[!htb]
\begin{minipage}{\columnwidth}
\resizebox{0.9\textwidth}{!}{
\centering

\tikzset{every picture/.style={line width=0.75pt}} 

\begin{tikzpicture}[x=0.75pt,y=0.75pt,yscale=-1,xscale=1]

\draw (261.08,275.53) node  {\includegraphics[width=222.99pt,height=67.81pt]{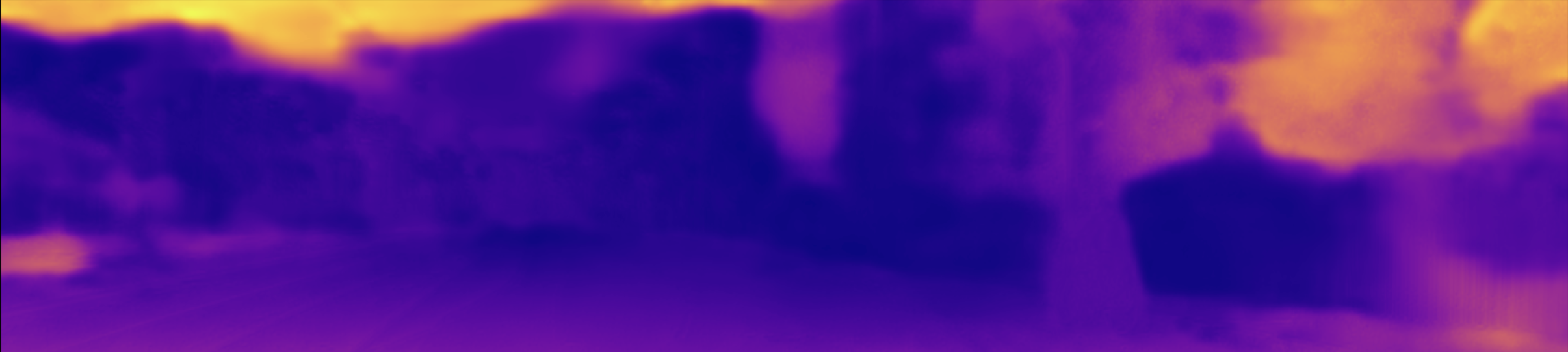}};
\draw (261.08,62.86) node  {\includegraphics[width=222.99pt,height=67.81pt]{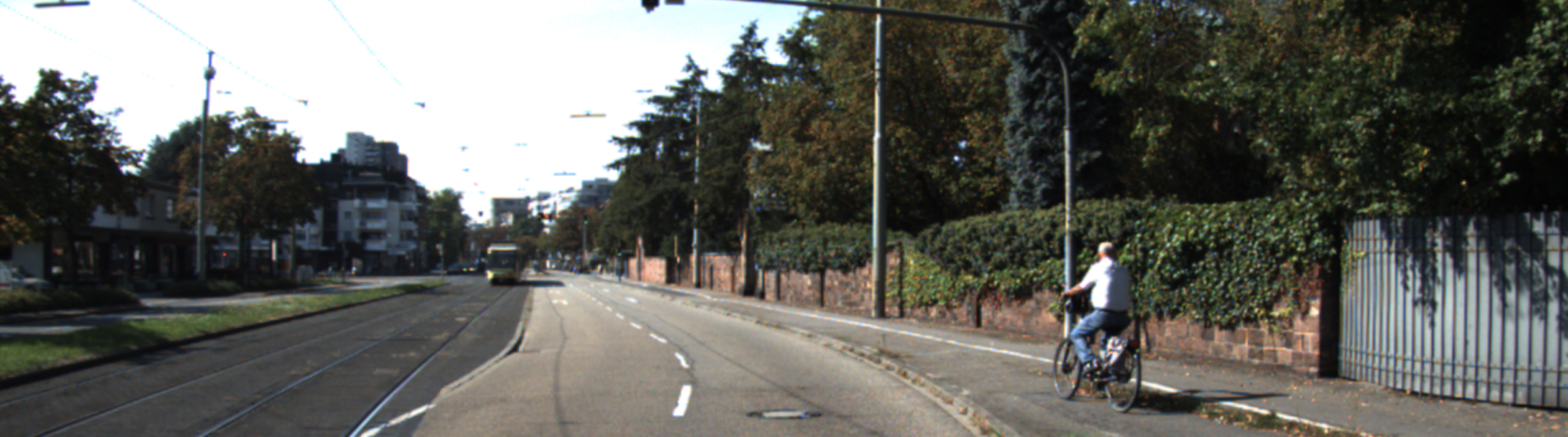}};
\draw (261.08,169.53) node  {\includegraphics[width=222.99pt,height=67.81pt]{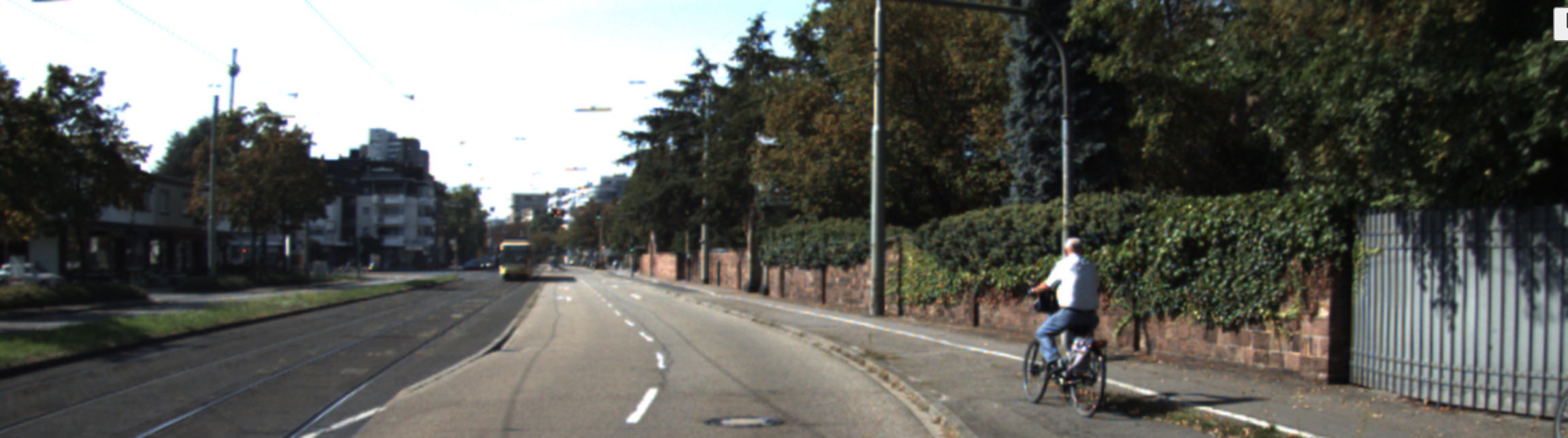}};
\draw   (112.42,17.65) -- (409.74,17.65) -- (409.74,108.07) -- (112.42,108.07) -- cycle ;
\draw   (112.42,230.32) -- (409.74,230.32) -- (409.74,320.74) -- (112.42,320.74) -- cycle ;
\draw   (112.42,124.32) -- (409.74,124.32) -- (409.74,214.74) -- (112.42,214.74) -- cycle ;

\draw (85.68,92.85) node [anchor=north west][inner sep=0.75pt]  [rotate=-270] [align=left] {Left Image};
\draw (85.68,303.03) node [anchor=north west][inner sep=0.75pt]  [rotate=-270] [align=left] {Prediction};
\draw (85.68,206.53) node [anchor=north west][inner sep=0.75pt]  [rotate=-270] [align=left] {Right Image};

\end{tikzpicture}

}
\end{minipage}
  \caption{\rev{On \textbf{top}: Left-view input image from the KITTI dataset rectified using Direct Stereo Rectification. On \textbf{middle}: Right-view input image from the KITTI dataset rectified using Direct Stereo Rectification. On \textbf{bottom:} Results of our model (CCNeXt) on this input pair.}}
  
  \label{fig:lim}

\end{figure}
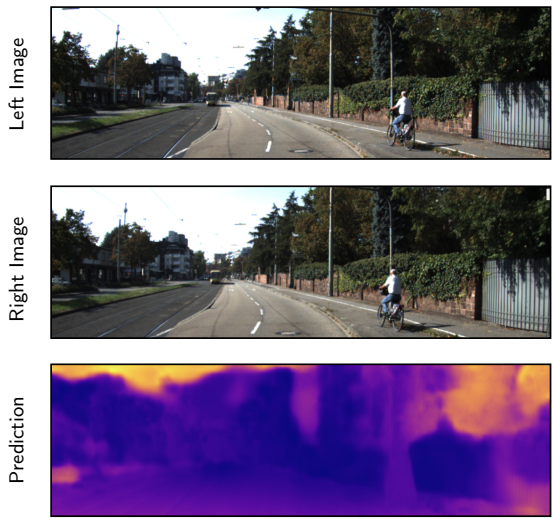

Finally, we need a pair of rectified images to do the depth and disparity matching. If all epipolar lines are parallel to the horizontal axis and corresponding points are in the same vertical coordinate, we satisfy the properties of a rectified image~\citep{loop1999computing}. To fix unrectified image pairs, online rectification methods have been extensively researched~\citep{wang2023practical,kumar2018multi,rehder2017online, LafioscaDirectStereoRectification},\rev{ but they were not extensively explored in our work. We conducted qualitative experiments with Direct Stereo Rectification~\citep{LafioscaDirectStereoRectification} on the raw KITTI data, which contains unrectified stereo pairs. An example is shown in \Cref{fig:lim}. The prediction reveals coarse errors introduced by inaccuracies in the online rectification step. These errors are particularly visible when comparing the orientation of vertical structures (e.g., poles in the scene center) across the two views, where imperfect alignment propagates into disparity estimation mistakes.}

\rev{
Beyond these specific limitations, we also highlight several developmental tendencies and open challenges for the field. 
First, integrating contrastive or self-distillation objectives with geometry-constrained attention may further enhance representation quality without increasing supervision requirements. 
Second, while CCNeXt demonstrates strong efficiency, future research should explore even lighter architectures to enable deployment on embedded platforms with strict latency and power budgets. 
Finally, robustness under adverse environmental factors such as extreme weather, nighttime, or sensor noise remains a critical challenge, where larger-scale datasets and adaptive learning strategies will be essential.}

\section{Conclusions}

This paper introduced CCNeXt, a novel convolutional network architecture designed for self-supervised stereo depth estimation. We enhanced the encoder module by incorporating a contemporary convolutional feature extractor coupled with a novel windowed epipolar cross-attention module. Additionally, we redesigned the widely employed Monodepth2 decoder to facilitate improved backpropagation and feature representations, resulting in enhanced estimations. \rev{Experiments demonstrate that CCNeXt not only achieves state-of-the-art results on KITTI but also establishes new benchmarks on DrivingStereo, outperforming prior methods in both the general split and the challenging weather-specific splits. 
This highlights the robustness of our approach under diverse conditions while being 10.18$\times$ faster than the current best model on the KITTI dataset.}

\section{Acknowledgments}
\label{acknowledgments}

The authors are thankful to the National Council for Scientific and Technological Development (grant \#304836/2022-2) for its financial support.

\bibliographystyle{model2-names}
\bibliography{paper}

\end{document}